\documentclass{article}

\usepackage{arxiv}
\usepackage[utf8]{inputenc} 
\usepackage[T1]{fontenc}    
\usepackage{natbib}
\renewcommand{\cite}{\citep}

\usepackage[caption=false,font=scriptsize,labelfont=sf,textfont=sf]{subfig}

\usepackage{bm}

\usepackage{mathptmx}                  

\usepackage{multirow}
\usepackage[dvipsnames]{xcolor}

\usepackage{here}

\usepackage{amsmath,amssymb,amsfonts}
\usepackage{array}
\usepackage{textcomp}
\usepackage{url}
\usepackage{verbatim}

\usepackage{graphicx}                

\usepackage{microtype}                 
\PassOptionsToPackage{warn}{textcomp}  
\usepackage{times}                     
\usepackage{booktabs}

\usepackage[hidelinks]{hyperref}
\hypersetup{
    colorlinks=true,
    linkcolor=NavyBlue,
    urlcolor=NavyBlue,
    citecolor=NavyBlue,
}

\title{Motion Transfer-Enhanced StyleGAN for \\Generating Diverse Macaque Facial Expressions}
\newif\ifuniqueAffiliation
\uniqueAffiliationfalse

\ifuniqueAffiliation 
\author{\href{https://orcid.org/0000-0002-0182-834X}{\includegraphics[scale=0.06]{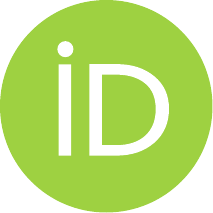}\hspace{1mm}Takuya~Igaue}\thanks{The codes and trained models are available at \ourgithub} \\
        National Institute of Advanced Industrial \\Science and Technology (AIST), Japan. \\
	\texttt{igaue@robot.t.u-tokyo.ac.jp} \\
	\And
	\href{https://orcid.org/0000-0002-2819-6039}{\includegraphics[scale=0.06]{orcid.pdf}\hspace{1mm}Catia~Correia-Caeiro} \\
	Human Biology \& Primate Cognition, \\Institute of Biology, \\Leipzig University, Germany. \\
	\texttt{catia\_caeiro@hotmail.com} \\
	\And
	\href{https://orcid.org/0000-0002-6180-1296}{\includegraphics[scale=0.06]{orcid.pdf}\hspace{1mm}Akito~Yoshida} \\
	Araya Inc., Japan. \\
	\texttt{yoshida\_akito@araya.org} \\
	\And
	\href{https://orcid.org/0000-0001-6502-655X}{\includegraphics[scale=0.06]{orcid.pdf}\hspace{1mm}Takako~Miyabe-Nishiwaki} \\
	Center for the Evolutionary Origins \\of Human Behavior (EHuB), \\Kyoto University, Japan. \\
	\texttt{miyabe.takako.2s@kyoto-u.ac.jp} \\
	\And
	\href{https://orcid.org/0000-0003-3866-3654}{\includegraphics[scale=0.06]{orcid.pdf}\hspace{1mm}Ryusuke Hayashi} \\
        National Institute of Advanced Industrial \\Science and Technology (AIST), Japan. \\
	\texttt{r-hayashi@aist.go.jp}
}
\else
        \usepackage{authblk}
        
        \newbox{\orcid}\sbox{\orcid}{\includegraphics[scale=0.06]{orcid.pdf}} 
        \author[1,2]{%
                \href{https://orcid.org/0000-0002-0182-834X}{\usebox{\orcid}}\hspace{1mm}Takuya~Igaue\thanks{\texttt{igaue@robot.t.u-tokyo.ac.jp}}%
        }
        \author[3]{%
                \href{https://orcid.org/0000-0002-2819-6039}{\usebox{\orcid}}\hspace{1mm}Catia~Correia-Caeiro\thanks{\texttt{catia\_caeiro@hotmail.com}}%
        }
        \author[4]{%
                \href{https://orcid.org/0000-0002-6180-1296}{\usebox{\orcid}}\hspace{1mm}Akito~Yoshida\thanks{\texttt{yoshida\_akito@araya.org}}%
        }
        \author[5]{%
                \href{https://orcid.org/0000-0001-6502-655X}{\usebox{\orcid}}\hspace{1mm}Takako~Miyabe-Nishiwaki\thanks{\texttt{miyabe.takako.2s@kyoto-u.ac.jp}}%
        }
        \author[1]{%
                \href{https://orcid.org/0000-0003-3866-3654}{\usebox{\orcid}}\hspace{1mm}Ryusuke~Hayashi\thanks{\texttt{r-hayashi@aist.go.jp} (Corresponding author)}%
        }
        \affil[1]{Human Informatics Research Institute, National Institute of Advanced Industrial Science and Technology (AIST), 1-1-1 Umezono, Tsukuba, Ibaraki, 305-8568, Japan.}
        \affil[2]{Graduate School of Engineering, The University of Tokyo, Japan.}
        \affil[3]{Human Biology \& Primate Cognition, Institute of Biology, Leipzig University, Germany.}
        \affil[4]{Araya Inc., Japan.}
        \affil[5]{Center for the Evolutionary Origins of Human Behavior (EHuB), Kyoto University, Japan.}
\fi

\renewcommand{\shorttitle}{\small \textit{Igaue et~al.} Motion Transfer-Enhanced StyleGAN for Generating Diverse Macaque Facial Expressions}

\hypersetup{
pdftitle={Motion Transfer-Enhanced StyleGAN for Generating Diverse Macaque Facial Expressions},
pdfsubject={eess.IV},
pdfauthor={Takuya~Igaue,Catia~Correia-Caeiro,Akito~Yoshida,Takako~Miyabe-Nishiwaki,Ryusuke Hayashi},
pdfkeywords={Macaque monkey, Action disentanglement, Facial expression transfer, StyleGAN2},
}

\begin{document}

\newcommand{\TI}{T.I.}
\newcommand{\AY}{A.Y.}
\newcommand{\CCC}{C.CC.}
\newcommand{\TMN}{T.MN.}
\newcommand{\RH}{R.H.}
\newcommand{\VNG}{Vanessa Nadine Gris}
\newcommand{\ourgithub}{\url{https://github.com/tigaue/maqface-stylegan2}}
\newcommand{\aist}{the National Institute of Advanced Industrial Science and Technology (AIST)}
\newcommand{\aistshort}{AIST}
\newcommand{\abci}{AI Bridging Cloud Infrastructure}
\newcommand{\abciurl}{\url{https://abci.ai/}}
\newcommand{\ehub}{EHuB}
\newcommand{\mtnaffiliation}{the Center for the Evolutionary Origins of Human Behavior (EHuB), Kyoto University}
\newcommand{\koshimaisland}{Koshima Island}
\newcommand{\koshimaprogram}{the Koshima Field Science Course and cooperation research program of the Wildlife Research Center, Kyoto University}
\newcommand{\acka}{the Japan Science and Technology Agency, Moonshot Research \& Development Program grant JPMJMS2012}
\newcommand{\ackb}{the National Institute of Information and Communications Technology (NICT) grant NICT 22301}
\newcommand{\ackc}{MEXT/JSPS KAKENHI Grant-in-Aid for Transformative Research Areas (A), Grant Number 24H02185}
\newcommand{\ackd}{Grant-in-Aid for Scientific Research (B), Grant Number 24K03241}

\newcommand{\baredteeth}{Bared-teeth}
\newcommand{\bark}{Bark}
\newcommand{\blink}{Blink}
\newcommand{\browraise}{Brow-raise}
\newcommand{\coo}{Coo}
\newcommand{\scream}{Scream}
\newcommand{\threat}{Threat}
\newcommand{\yawn}{Yawn}
\newcommand{\lipsmack}{Lip-smack}
\newcommand{\chewing}{Chewing}
\newcommand{\lookup}{Look-up}
\newcommand{\lookdown}{Look-down}
\newcommand{\lookleft}{Look-left}
\newcommand{\lookright}{Look-right}
\newcommand{\tongueprotrude}{Tongue-protrusion}
\newcommand{\tongueshow}{Tongue-show}

\newcommand{\pleasecheck}[1]{\textcolor{black}{#1}}

%
%
\newcommand*{\figref}[2][]{%
\hyperref[{fig:#2}]{%
  Fig.~\ref*{fig:#2}%
  \ifx\\#1\\%
  \else
    \,(#1)%
  \fi
}%
}
\newcommand*{\tabref}[2][]{%
\hyperref[{tab:#2}]{%
  Table~\ref*{tab:#2}%
  \ifx\\#1\\%
  \else
    \,(#1)%
  \fi
}%
}
\newcommand{\eq}[1]{\begin{align}#1\end{align}}

\maketitle
\begin{figure*}[t]
  \centering
  \includegraphics[width=\textwidth, alt={Results of manipulating macaque face images using the StyleGAN2 model trained using the proposed method.}]{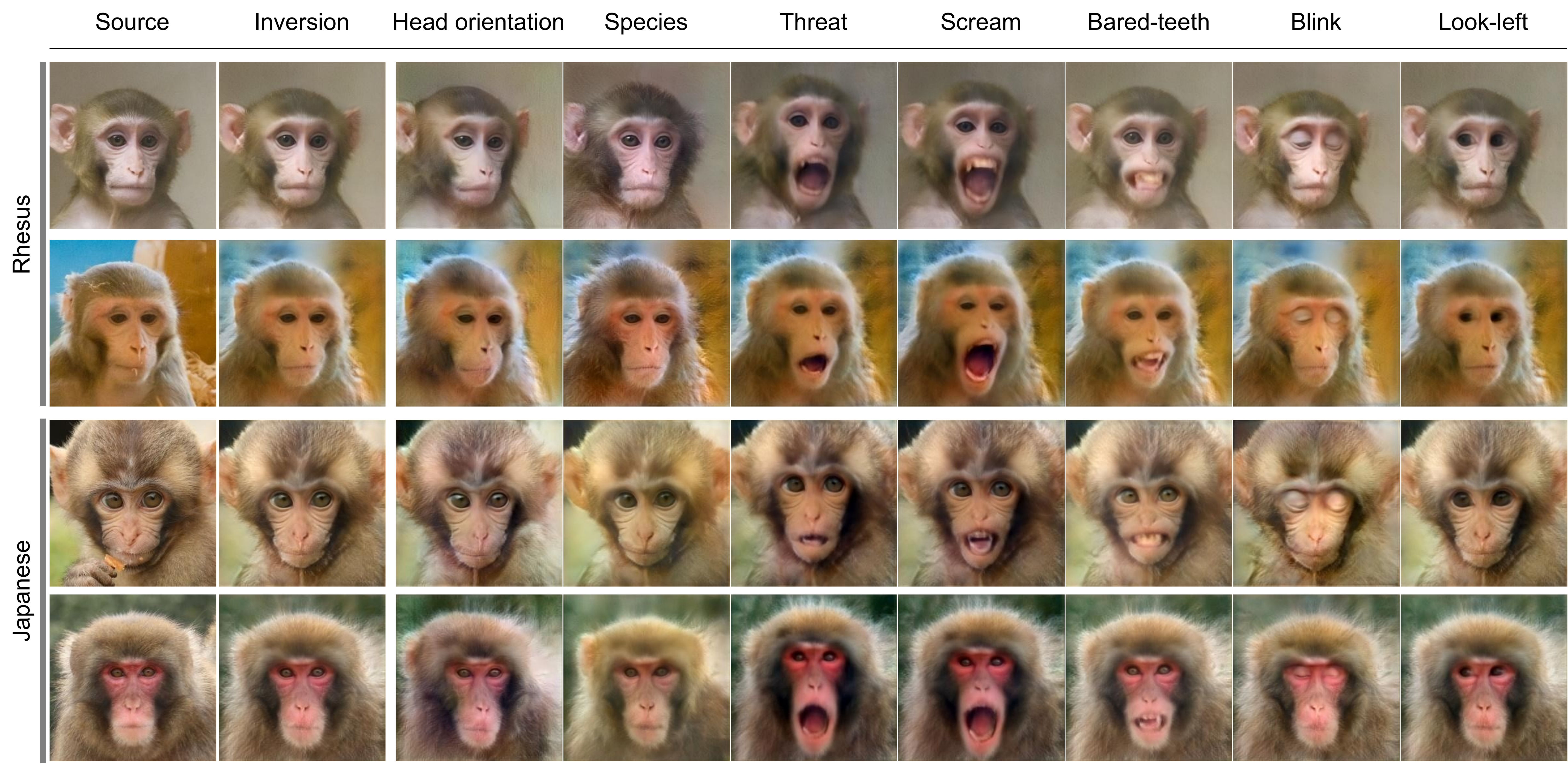}
  \caption{Results of manipulating macaque face images using the StyleGAN2 model trained using the proposed method. The images labeled Inversion were generated from the estimated latent codes of the source images using the trained model and demonstrate successful latent learning. The trained model also acquired a well-disentangled latent representation, which enabled the generation of photorealistic macaque face images with various semantic edits, such as adjusting the head orientation to the right (third column labeled Head orientation) and alternating the appearance between two species (i.e., Japanese macaque and Rhesus macaque), labeled Species. Additionally, the model can generate various characteristic macaque facial expressions, including \threat{}, \scream{}, and \baredteeth{}, in addition to eye movements such as \blink{} and \lookleft{}.}
  \label{fig:teaser}
\end{figure*}

\begin{abstract}
Generating animal faces using generative AI techniques is challenging because the available training images are limited both in quantity and variation, particularly for facial expressions across individuals. In this study, we focus on macaque monkeys, widely studied in systems neuroscience and evolutionary research, and propose a method to generate their facial expressions using a style-based generative image model (i.e., StyleGAN2). To address data limitations, we implemented: 1) data augmentation by synthesizing new facial expression images using a motion transfer to animate still images with computer graphics, 2) sample selection based on the latent representation of macaque faces from an initially trained StyleGAN2 model to ensure the variation and uniform sampling in training dataset, and 3) loss function refinement to ensure the accurate reproduction of subtle movements, such as eye movements. Our results demonstrate that the proposed method enables the generation of diverse facial expressions for multiple macaque individuals, outperforming models trained solely on original still images. Additionally, we show that our model~\footnote{The codes and trained models are available at \ourgithub} is effective for style-based image editing, where specific style parameters correspond to distinct facial movements. These findings underscore the model’s potential for disentangling motion components as style parameters, providing a valuable tool for research on macaque facial expressions.
\end{abstract}

\keywords{Macaque monkey \and Action disentanglement \and Facial expression transfer \and StyleGAN2}


\section{Introduction}

The presentation and analysis of animal face images have been used 
in behavioral and system neuroscience research to explore the animals' cognitive functions 
related to facial recognition and social behavior. 
Among various experimental animals, the macaque monkey is of particular importance as a non-human primate 
because of its biological, anatomical, and physiological similarities to humans, including the musculoskeletal structure, 
reproduction, and immunity~\cite{Waller2008}. 

Three-dimensional (3D) rendering techniques in computer simulation are frequently used to create images and videos of animal faces, including those of macaques~\cite{Zuffi2017}. However, generating realistic animal faces via computer graphics (CG) remains challenging 
because accurately rendering light reflection, absorption, and diffusion in animal skin and hair, 
in addition to depicting naturalistic facial movements, 
requires the precise physical modeling of animals' facial structures and motion capturing, 
and thus incurs in high computational costs~\cite{Egger2020}. Recently, StyleGAN2~\cite{Karras2020,Karras2020a}, which embeds images in a well-disentangled latent space represented by style parameters, has generated realistic images of human faces from vast training images without the need for detailed 3D modeling. This disentanglement ability makes the application of StyleGAN2 wide, ranging from editing facial attributes, such as eye size, hair color, and sex~\cite{Shen2020,Wu2021} to emotion estimation~\cite{Kiyokawa2024}. The analysis of StyleSpace, which is the space of channel-wise style parameters for each layer of the generator, enables a particular channel to correspond to particular motions of the human face~\cite{Wu2021}. This correspondence between style parameters and facial motions observed in the StyleGAN2 model for human faces is expected to extend to other animal faces, particularly non-human primates because of their similar musculoskeletal structure. Such correspondence is potentially valuable for analyzing facial movements in animals, similar to the action unit (AU) analysis of primate faces~\cite{Darwin2019,correia2021extending,Parr2010}, and for mapping facial components across species~\cite{Waller2020}, which enable comparisons between humans and other primates, thereby offering insights into evolutionary mechanisms of communication and emotion.

Generating realistic animal faces remains challenging with StyleGAN2 because the available training images are limited in their variation. This is true for some species of macaques, which despite being widely used for research purposes, have a limited diversity of images readily available in datasets, particularly in terms of different facial expressions. Importantly, primates rely heavily on facial expressions for communication in social situations. For instance, macaques can recognize facial expressions in conspecifics~\cite{Parr2009} and use information from these visual displays to predict social interactions~\cite{Waller2016}. The importance of faces in primates' social interaction is supported by their sensitivity to even the slightest deviations from typical faces, a psychological phenomenon known as the uncanny valley effect ~\cite{Carp2022,Siebert2020,Steckenfinger2009,Wilson2020,IGAUE2023107811}. Additionally, several brain areas, referred to as face-selective patches, are selectively activated when humans and macaques detect face stimuli~\cite{Dureux2023,Hesse2020}.
Therefore, there is great interest in overcoming the limitations of the facial expression editability of current image generation techniques to allow for more advanced face image analyses and behavioral experiments that use more photorealistic facial appearances and corresponding interactions with generated facial images. 

To address this challenge, the training image dataset must incorporate individual variation and more realistic facial expressions. The objective of this study is to propose a method for training the StyleGAN2 model with expression-specific data augmentation of macaque faces and to analyze the resulting latent semanticity to explore the future application of facial expression analysis in non-human primates.

The contributions of this paper are as follows:
1) We use a motion transfer technique for facial-expression-specific data augmentation in StyleGAN2 training. 
2) We train a motion transfer model using our original movie dataset, 
which captures the movements of several facial expressions in Japanese and Rhesus macaques. 
3) We train the StyleGAN2 model using augmented facial expression images, use a sampling method to prevent latent shrinkage around rare expressions, 
and use our own loss function to replicate the subtle image changes in eye movements. 
4) To evaluate the disentanglement of the trained StyleGAN2 model, 
we explore the latent space and determine that several style parameters correspond to specific facial movements, 
such as those of the mouth and eyes.
\section{Related Work}
\subsection{Visual models of macaque face in previous animal research}
\subsubsection{Three-dimensional computer graphic models}


In several studies, researchers have created 3D models of macaque faces, 
whose expression is controllable through a user interface for research purposes~\cite{Steckenfinger2009}. 
In some studies, researchers have edited macaque faces to generate experimental stimuli by modifying texture data for a 3D polygon model created by 3D CG software~\cite{Carp2022} or creating a 3D model by 3D scanning the skeletal structure~\cite{Wilson2020}. Siebert and colleagues manipulated macaque face surface models, created from MRI images with motion capture results obtained from acting macaque faces~\cite{Siebert2020}. 
MF3D~\cite{Murphy2019} is a precise 3D model of macaque heads based on anatomical knowledge of macaques (particularly the Rhesus macaque). It partially incorporates some AUs (see Section~\ref{subsec:au} for details) in its movement. The quality of face images rendered by MF3D is superior to that of the other models, which allows for the control and replication of realistic motion patterns for both facial expressions and head orientation while maintaining the animal's identity~\cite{Taubert2020}. However, a drawback of MF3D is that the rendered images are not photorealistic, and the appearance of individual faces is manipulated by only a few parameters that determine the musculoskeletal structure of the model and textures of skin and hair, which limits its ability to reproduce the precise facial appearance of specific individuals. Given the face-specific sensitivity of macaques, the photorealistic quality of macaque face images is critical for investigating the animal's ecologically valid functions~\cite{Wilson2020}. To date, no previous studies based on 3D CG modeling approaches open for research purposes have achieved the generation of photorealistic facial images of macaques with variations in realistic facial expressions.

\subsubsection{Photometric models}

An alternative approach to 3D CG modeling is to use machine learning techniques to generate facial images from image datasets. 
Numerous methods, including ProgressiveGAN~\cite{Karras2018}, StyleGANs~\cite{Karras2021a,Karras2020a,Karras2021}, and diffusion models~\cite{Ho2020}, have been proposed to generate individual variations of photorealistic face images. In particular, StyleGAN2 is a widely used method with strong editability. 
Encoding methods, such as pixel2style2pixel (pSp)~\cite{Richardson2020}, e4e~\cite{Tov2021}, and ReStyle-encoder~\cite{alaluf2021restyle}, have also been proposed to estimate the style parameters of a trained StyleGAN2 model from given images using feedforward neural networks. The combination of StyleGAN2 and its encoding models allows for the editing of source images along a certain manipulation direction in the disentangled latent spaces.
Furthermore, when trained on a diverse set of human individuals, StyleGAN2 effectively disentangles facial motions~\cite{Tov2021,Wu2021}, and its disentangled latent representation is also available for facial expression analysis\cite{Kiyokawa2024}. However, StyleGAN2 and its encoder struggle to achieve desirable image generation and facial expression editing when training samples are limited. 
As a general approach for training StyleGAN2 using limited data, data augmentation techniques have been specifically designed to prevent overfitting and leakage to the discriminator (e.g. StyleGAN2-ADA~\cite{Karras2020} and DiffAugment~\cite{Zhao2020}). This scheme enables the StyleGAN2 model to avoid mode collapse and achieve better image generation quality from the estimated latent codes of a given image, which is a process referred to as “inversion,” while also improving editability. Although these methods have proven effective in the domain of non-primate faces, such as cats, pandas, and dogs~\cite{Zhao2020} and alleviate some challenges in image training, the data augmentation techniques used in the previous studies rely on standard image perturbations (e.g., translation and color jitters), which do not help to increase variations in individual faces or facial expressions that StyleGAN2 can generate. 

By contrast, motion transfer methods, such as the first-order motion model for image animation (FOMM)~\cite{Siarohin2019}, motion representations for articulated animation (MRAA)~\cite{Siarohin2021}, and thin-plate spline motion model (TPSMM)~\cite{Zhao2022}, enable the animation of images by transferring motion from driving videos to source images. In the context of face animation, this technique can apply facial expressions from a video to still images of different individuals, thereby generating highly natural videos of facial expressions on those individuals. Although motion transfer methods lack the ability to edit individual facial features or movements without using specific source images and driving videos, they have potential for use in data augmentation to increase variations in facial expressions. 

Human facial research has traditionally used six or seven labels for holistic facial expressions~\cite{Ekman1971} to study facial emotion perception~\cite{Adolphs2006}. Likewise, macaques, which use facial expressions for social communication, exhibit several species-typical holistic expressions~\cite{Maestripieri1997,Maestripieri1997a}. Because of the similarity in musculoskeletal structures among individuals of the same species, motion transfer methods that process changes in the motions and textures at distinct facial regions are expected to realistically reproduce facial movements across different macaques particularly if the expressions are species-typical.

\subsection{Disentanglement of facial motion components in macaque}\label{subsec:au}

The AU is the minimal component used to identify facial movement linked to underlying musculature within the Facial Action Coding System (FACS)~\cite{Waller2020}. 
Previous machine learning approaches for automating FACS analysis used landmark detection or the latent representation of StyleGAN2 with annotation information of the AUs of human faces~\cite{Baltrusaitis2015,yin2024fg}. To date, only one automated system has been developed for the tracking of some AUs in macaques~\cite{Morozov2021}. However, the use of an image-generating model to analyze facial motion components has not yet been examined for macaques. StyleGAN2 has demonstrated the ability to disentangle human facial features in its latent space without the need for annotation information for image editing and has been used for AU detection. Therefore, it is likely that the same framework could also disentangle macaque facial features and motions. However, this potential application has not been explored. 

\section{Methods}

\subsection{Overview of the training scheme using StyleGAN2}

\begin{figure*}[t]
  \centering
  \subfloat[][Overview of the motion transfer-enhanced StyleGAN2 training and style-based image editing.]
    {\includegraphics[clip,width=0.6\linewidth]{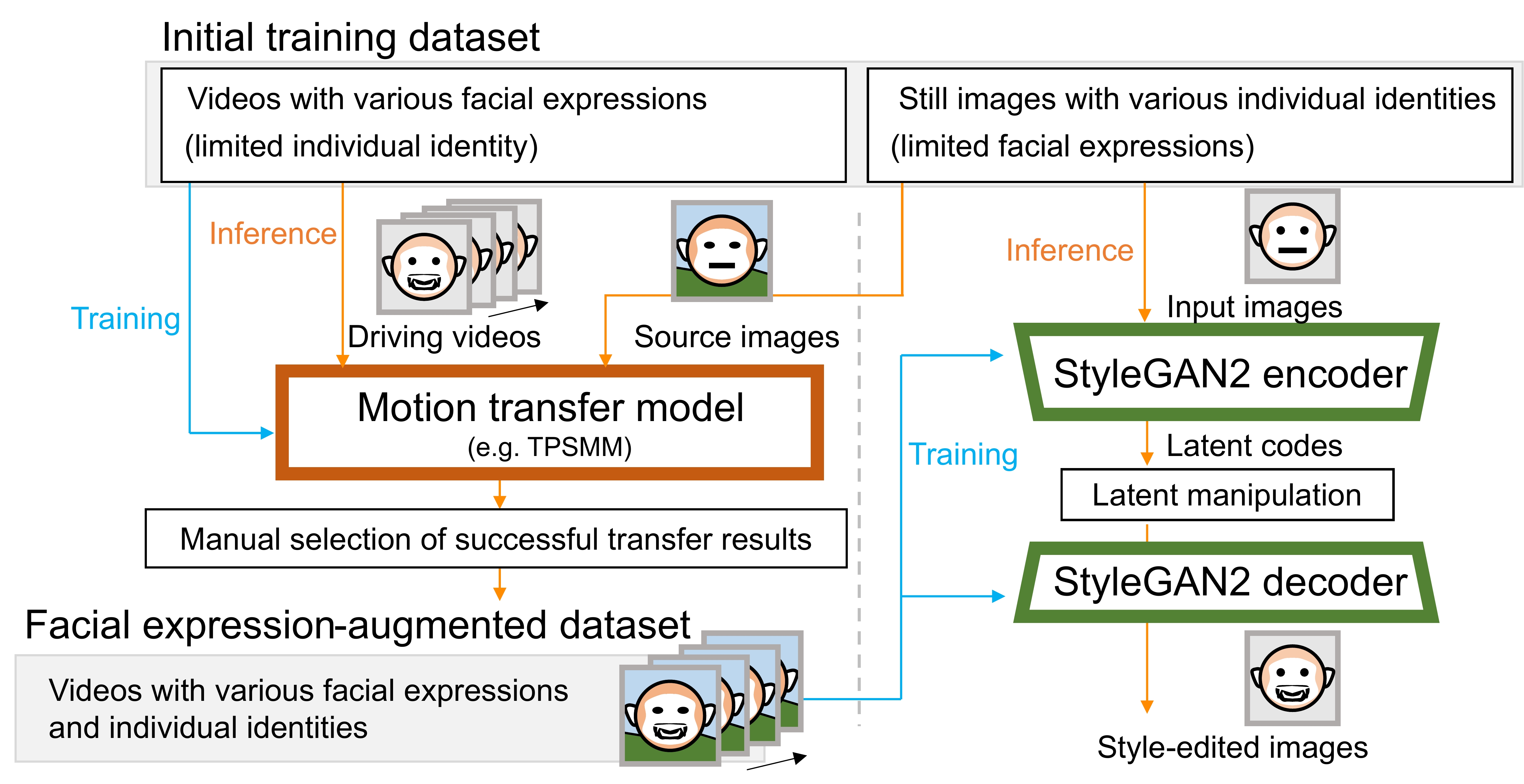}}\\
  \subfloat[][Workflow of the proposed StyleGAN2 training.]
    {\includegraphics[clip,width=0.8\linewidth]{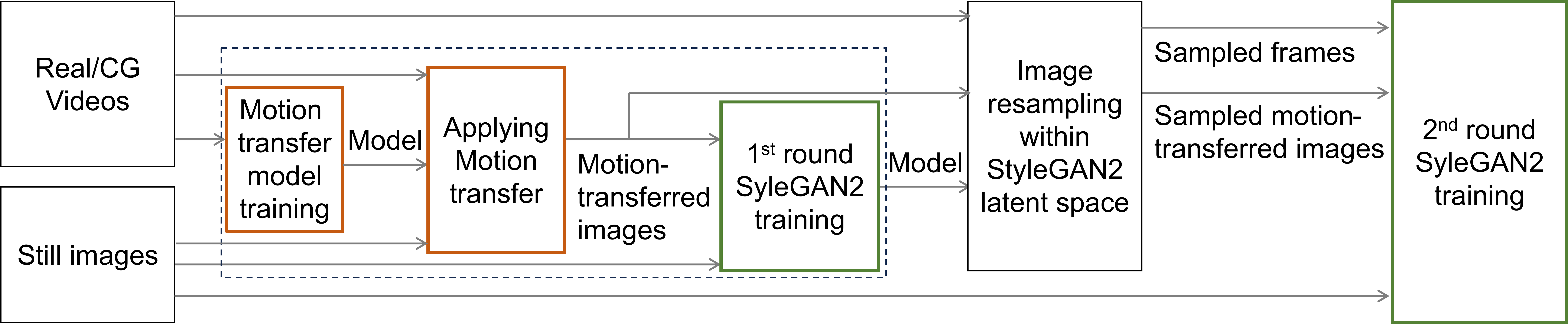}}
  \caption{Overview of our proposed method for generating diverse macaque facial expressions using the StyleGAN2 model. We incorporated facial expression data augmentation via motion transfer techniques, using driving videos created by realistic CG models (MF3D) to systematically transfer labeled facial expressions to still images of diverse macaque individuals. This method allowed us to expand the training dataset in terms of both quantity and expression variations. Additionally, in the second round of training, we applied a specialized loss function to enhance the inversion quality of subtle facial movements, particularly around the eyes and selected our training images to correct for biases in the neutral facial expressions and ensure greater diversity.}
  \label{fig:method-overview}
\end{figure*}

In this study, we developed a new method for generating photorealistic faces with wide variations of individual identity and facial expression repertories in two macaque species (Japanese macaques, \textit{Macaca} \textit{fuscata}, and Rhesus macaques, \textit{Macaca} \textit{mulatta}) based on StyleGAN2 with dedicated data augmentation and a loss function. An overview of this method is shown in \figref{method-overview}. 
We used publicly available macaque face images for training StyleGAN2, which predominantly have neutral expressions and lack variation in facial expressions. To expand the training dataset in terms of expression, we applied a motion transfer technique 
using videos of a limited number of macaque individuals that displayed various facial expressions to still images that mostly showed neutral expressions, but represented diverse identities.

To achieve this data augmentation, we first created an original video dataset for training the motion transfer model for macaque face image animation. The datasets included computer graphically generated videos, 
which are categorized into several macaque-typical expressions, in addition to videos recorded from real macaques (see Section~\ref{sec:method-prepvid} for details).

After we successfully trained the motion transfer model with macaque video datasets, the model was able to transfer several macaque-typical facial expression movements in driving videos generated by the CG videos to arbitrary still macaque face images. We chose CG videos over real macaque videos as driving videos because the head orientations were perfectly aligned across different videos, and the facial expression movements were idealized and labeled with their respective category names, which allowed us to systematically transfer facial expressions to other still face images of diverse macaque individuals to expand the training datasets for StyleGAN2 with annotated labels of the synthesized expressions. 

We observed that training StyleGAN2 using each frame from motion-transferred videos, along with the source still image datasets (i.e., the first training), was not sufficient to generate macaque facial images with a wide range of variations in expressions. This issue arose because the driving videos predominantly displayed neutral expressions, which caused the model to become overly specialized for neutral faces. To address this issue, we further sampled dissimilar images from the dataset by leveraging the latent distribution from the first training of the StyleGAN2 model. 
In the second training stage, we included only images distant in the latent space. Even with this unbiased sampling, the model still struggled to generate subtle movements, such as eye movements, which are rare in the dataset and exhibit small pixel-level changes compared with overall image diversity. 
Therefore, to improve the quality of generated eye movements, we introduced a custom loss function in the second round of training, with additional weight applied to reconstruction errors in the eye region. 

We evaluated the StyleGAN2 model trained using our method in terms of the variety of facial expressions it could generate, its image editing capabilities, and disentanglement qualities over some annotated facial features.

\subsection{Preparation of the macaque face dataset}

\subsubsection{Still image dataset}\label{sec:method-prepstil}

To create a training face image dataset with a wide variety of macaque individuals,  
we obtained several image datasets of Japanese and Rhesus macaques provided for research purposes from the following four archives: 
1) images labeled \textit{n3} in \textit{10\_monkey\_species}~\footnote{\url{https://www.kaggle.com/datasets/slothkong/10-monkey-species}}, which the creator of the dataset originally downloaded from the internet 
(152 images of Japanese macaques); 
2) \textit{Macaque}\textit{\_Faces\_9862586}~\footnote{\url{https://figshare.com/articles/dataset/Macaque_Faces/9862586}} 
recorded at the Medical Research Council's Centre for Macaques, Salisbury, UK~\cite{Witham2019} (3895 images of Rhesus macaques); 
3) Visiome PrimFace~\footnote{\url{https://visiome.neuroinf.jp/database/list/6948}} recorded at several macaque research facilities in Japan 
(316 images of Japanese and 301 images of Rhesus macaques); and 
4) our original images recorded at~\mtnaffiliation{} (2,153 images of Japanese macaques). 
From the downloaded images, we cropped the face area to 
256 $\times$ 256 pixels using the image analysis tool 
InsightFace~\footnote{\url{https://insightface.ai/}}, 
which includes a precise 106 two-dimensional landmark detection library 
for the human face. 
We verified that this library also works for macaque faces. 
After detecting the landmarks, we warped the eye and mouth positions to match the designated locations in pSp~\cite{Richardson2020}, which is an encoding model of StyleGAN2, for human face images. 
After cropping the images, we split the dataset into approximately 
95\% (5638 images) training images and 5\% test images (300 images). 
We performed this split before motion transfer data augmentation 
to ensure consistency in training/test individuals throughout this study.

\subsubsection{Real video dataset}\label{sec:method-prepvid}

\RH{} recorded 115 video clips of 15 Japanese macaques housed at \aist{} as they sat in a primate restraint chair with free head movement using cameras (ZV-E10L, Sony, and iPhone14 Pro Max, Apple Inc.). Additionally, we sourced 22 publicly available videos of Japanese macaques from the wild, which included approximately 30 individuals. We also recorded 76 video clips of Japanese and Rhesus macaques housed at \ehub{} in an outdoor enclosure (recorded by \CCC{} and \TMN{} using Panasonic HC-V480MS and HC-WX970M-K), in addition to provisioned wild Japanese macaques from two troops of approximately 100 individuals in total that live on \koshimaisland{} (recorded by \CCC{} and \VNG{} during \koshimaprogram{} using GoPro HERO6Black and HERO7Black, and Panasonic HC-WX970M-K). 
All 213 videos ranged from 1 second to 26 minutes in length, totaling 12 hours and 45 minutes, with an average duration of 3 minutes and 36 seconds.

Subsequently, we cropped the facial region from the raw video frames using the library of InsightFace. To track continuous facial movements, we applied a bounding box intersection to the facial area, similar to the approach used in a previous study for training FOMM~\cite{Siarohin2019} on the human face image dataset (VoxCeleb datasets~\cite{Nagraniy2017}) with a threshold of 0.2. We used a lower threshold than the default value of 0.5 to reduce interruptions in video cropping caused by missed face detection. Additionally, to ensure that the StyleGAN model learned facial movements rather than head movements, we set the rule that the first frame of the video clips had to show an estimated head orientation within 15 degrees from the center using the pose parameter output by InsightFace. As a result, we obtained 5,152 cropped videos showing various facial movements of real macaques with approximately centered head orientations.

\subsubsection{CG video dataset}

To train StyleGAN2 with images containing labeled facial expressions, we used CG video clips created by MF3D~\footnote{\url{https://github.com/Phenomenal-Cat/MF3D-Tools}}~\cite{Murphy2019}. These included 33 videos depicting 11 macaque-typical facial expressions (i.e., \baredteeth{}, \bark{}, \blink{}, \browraise{}, \chewing{}, \coo{}, \lipsmack{}, \scream{}, \threat{}, \tongueprotrude{}, \yawn{}) viewed at yaw angles of $[-30, 0, 30]$ degrees from the center, in addition to six videos showing head rotation movements in the yaw direction. We cropped the video clips using the same procedure as that for real videos, but without applying bounding box intersection criteria. This step aimed to capture sequential motion from a neutral face to the target facial expression, with the head position stable throughout video clips and resulted in a total of 40 cropped videos. 

\subsection{Animating still images with facial expressions using a motion transfer technique}

We took advantage of motion transfer techniques, which can animate a source image with the facial expressions from driving videos while keeping its original texture, to create a macaque face image dataset with a wide range of identities and facial expressions for subsequent StyleGAN2 training. 

We used the TPSMM model~\cite{Zhao2022}, which enables accurate motion transfer by separately matching texture and motion flow between trained and generated images using a few key points to warp a nonlinear plane. This approach reduces the need for extensive hyperparameter tuning. Using CG videos as driving videos offers the advantage of systematically applying facial expressions to still images while faithfully replicating the facial movements of the real macaque. To capture as many variations of facial movement as possible, we trained TPSMM on both real and CG video crops, with the datasets consisting of 95\% real videos (4,894 videos, each ranging from 1 to 5 seconds) and 40 CG videos (less than 2 seconds each). We used the remaining 5\% of the real videos to validate the trained TPSMM model.

We trained a TPSMM model on a workstation (CPU, Intel Xeon Gold 6326 Processor x2; GPU, NVIDIA Tesla A100 PCIe (80 GB, HBM2e) x2). For hyperparameter tuning, we changed the number of thin-plate splines ($K$) between [10, 20, 30, 40, 50] using 75 training iterations and found that accuracy plateaued at 30 splines. Then we increased the number of iterations from 75 to 150 to achieve the final checkpoint. Quantitative parameter tuning results and sample motion transfer outputs are provided in \pleasecheck{Table~A3 and Fig.~A3} in the supplementary material, respectively.

After training the TPSMM model, we transferred facial expressions from the driving videos to the still image dataset. For the repertoire of facial expressions used in this data augmentation, we chose 11 facial expressions from the example MF3D CG videos. Additionally, we manually chose several eye movements and a tongue movement from the real videos. In total, we selected 16 facial expressions, which we referred to as \baredteeth{}, \bark{}, \blink{}, \browraise{}, \chewing{}, \coo{}, \lipsmack{}, \scream{}, \threat{}, \tongueprotrude{}, \yawn{}, \lookup{}, \lookdown{}, \lookleft{}, \lookright{}, and \tongueshow{}. To apply 11 of these facial expression types from the CG videos to the real still images, we adjusted the head orientation of the driving videos from $[-30, 0, 30]$ to match the source images based on the head orientation estimated by InsightFace. We used the relative motion transfer mode of TPSMM to achieve high-quality motion transfer results. The motion transfer quality strongly depends on the difference between the first frame of the driving video and the source image, for example, the results are degraded if the eyes are closed in the source image but open in the initial frame of the driving video. Therefore, we manually chose high-quality motion transfer results, particularly based on the quality of eye and teeth images, for the subsequent process. Then we split the final synthesized video clips into training and test videos, keeping the consistent training/test split labels of the still images from Section~\ref{sec:method-prepstil}, that is, the final video clips were synthesized from 353 training and 20 test source images using the trained TPSMM model. The basic dataset for training StyleGAN2 included both still images and video frames of their synthesized versions after motion transfer.

\subsection{Training of StyleGAN2}
\subsubsection{Codes and computing platforms}

To generate and edit the macaque face images, we used StyleGAN2, for which sophisticated generator and encoder neural network implementations are publicly available. To prevent mode collapse while achieving high image generation quality via data augmentation of relatively small image samples, we used StyleGAN2-ADA~\cite{Karras2020} for generator training. For encoder training, we used the ReStyle framework~\cite{alaluf2021restyle}, which iteratively improves inversion quality to generate more realistic images with the pSp encoder architecture and learning rules as its backbone. 

We used the official TensorFlow implementation of StyleGAN2-ADA provided in its GitHub repository~\footnote{\url{https://github.com/NVlabs/stylegan2-ada}}. 
We conducted training on a cloud computing server (CPU, Intel Xeon Gold 6148 Processor x2; GPU, NVIDIA Tesla V100 SXM2 (16 GB HBM2) x4) provided by \abci{}~\footnote{\abciurl} of \aistshort{}. 
We trained the encoding model using the ReStyle framework on 
another workstation (CPU, Intel Xeon Processor E5-2687W v4; GPU, NVIDIA GeForce RTX 3090 (24 GB)).

\subsubsection{Image dataset for the first round of training}

To capture a wide range of facial expression variations, we divided our StyleGAN2 model training procedure into two training steps. The first round of training focused on acquiring a macaque face generation model, which we then used to validate the similarity of macaque face images for sampling distinct images from the original datasets for the second round of training. 
\pleasecheck{The details of the dataset for the first round of training are described in the supplementary material~B1.}

\subsubsection{Image dataset for the second round of training: weighted random sampling to mitigate bias in the image dataset}

Having models trained on datasets with mostly neutral faces reduces the inversion quality for the modeling of facial expression variations. To address this issue, we added face images extracted from real videos to the dataset while eliminating redundant face images using the following sampling procedure.

To ensure greater variation in the training dataset for the second round of training, we sampled image frames from the entire set of real videos based on their distributions in the latent representation space of the StyleGAN2 model from the first round of training. First, we calculated the outputs of the trained encoder for all frames. Then, we randomly sampled images in this latent space with a weighting system based on the L2 distance from already sampled images, similar to the initial cluster determination of \textit{k}-means++~\cite{Arthur1996}, until the variety of sampled images was maximized. To quantify the diversity of the sampled images, we estimated entropy~\footnote{\url{https://github.com/gregversteeg/NPEET}} and tested different sample sizes [500, 1000, 2000, 3000, 4000, 5000, 10000] from all 450,218 frames in the real videos after face area cropping. We computed estimated entropy using jackknife resampling~\cite{Burnham1983} of 300 sub-samples. This step confirmed that 2,000 samples optimized the variety of images using the L2-weighted sampling method. 

To ensure a more balanced sampling of facial expressions, we selected 130 frames that displayed distinct facial expressions using the same weighted sampling procedure from the driving videos of 16 facial expression types for motion transfer. This included 100 frames from CG videos of 11 facial expressions; 10 frames from real videos featuring \lookup{}, \lookdown{}, and \tongueshow{} videos and 20 frames from real videos featuring \lookleft{} and \lookright{} movement videos. Only the motion-transferred synthetic images corresponding to these 130 selected frames were included in the second round of training. Consequently, we obtained 53,528 training images and 2,900 test images for the second round of training.

\subsubsection{Loss design for improving inversion quality around the eye area}

Although the default implementation of the ReStyle framework generates high-quality inversion images through its iterative improvement process, 
it is not sufficient to accurately replicate eye movements even after motion transfer data augmentation. 
This limitation arises because the sclera (white area of the eyes) in macaques 
is smaller than that of humans, and often not visible at all, 
which makes changes around the eye region quite subtle. 
To enhance the fidelity of inversion related to eye movements, 
we introduced an additional L2 loss specifically for the eye region alongside the standard L2 loss of the entire images, 
LPIPS loss~\cite{Zhang2018} (a variation of perceptual loss used to ensure natural image generation), 
and a similarity loss ($\mathcal{L}_{sim}$) to encourage the generation of diverse facial identities. 
The L2 loss for the eye region is calculated by masking the predetermined regions 
(horizontally, 1/4 to 3/4 of the image width, and vertically, 1/4 to 1/2 of the image height).
The following function defines the L2 loss for the eyes:

\eq{
  \mathcal{L}_{2_{eye}}(x) = \lambda_{l2_{eye}} \frac{A}{\sum \mathbf{M}} || (\mathbf{I} - \hat{\mathbf{I}}) \circ \mathbf{M}||^2,
}
where $\mathbf{I}$ is the inverted image, $\hat{\mathbf{I}}$ is the training image, 
$\mathbf{M}$ is the mask image, and $A$ is the area of the image. 
We explored the effect of using different values of $\lambda_{l2_{eye}}$ during training on inversion results, as shown in the Result section~\ref{subsec:result-inversion}. 
We set the weight parameters for the losses from the original ReStyle implementation to the default values of 
$\lambda_{l2}=1.0$, $\lambda_{lpips}=0.8$, and $\lambda_{sim}=0.5$.

We initially trained the generator using StyleGAN2-ADA, 
followed by encoder training within the ReStyle framework, fixing the generator parameters until the loss for test images reached the plateau (approximately 36,500 steps). 
Then we jointly trained both the generator and encoder for more than 300,000 steps. 
The similarity loss $\mathcal{L}_{sim}$ was calculated using the ResNet-50 model pretrained using MOCOv2 images, 
which is a commonly used approach for evaluating the similarity of objects other than human faces, 
and applied directly in this macaque face study to prevent the models from generating similar macaque faces.

\begin{table*}[t]
  \centering
  \caption{Positive and negative categories for mouth opening and eye closing motions}
  \label{tab:posneg_def}
  \begin{tabular}{l p{5.5cm} p{7.5cm}}
   \toprule
    & Positive motion & Negative motion \\
   \midrule
   Mouth opening & \baredteeth{}, \bark{}, \scream{}, \threat{}, \yawn{} & \blink{}, \browraise{}, \lipsmack{}, \lookup{}, \lookdown{}, \lookleft{}, \lookright{} \\
   Eye closing & \blink{}, \lookdown{} & \baredteeth{}, \bark{}, \browraise{}, \chewing{}, \lipsmack{}, \scream{}, \threat{} \\
   \bottomrule
  \end{tabular}
\end{table*}



\subsection{Style-based image editing using annotation information}

As an image manipulation technique to demonstrate the disentanglement of the latent representations learned by the model, we used annotation information from the dataset and applied the InterFaceGAN~\cite{Shen2022} procedure. This method allows us to extract the latent representation corresponding to a specific attribute of interest. If the attribute is well disentangled in the latent space, adding the extracted latent vector to the latent representation of source images edits the generated images, thereby enhancing the focused attribute. 
InterfaceGAN estimates the editing direction as a perpendicular vector of the classification boundary between positive and negative sample groups. To calculate the editing directions of macaque facial expressions, we labeled the target expressions defined in MF3D as positive and the corresponding neutral expressions as negative. Our training framework reduced the annotation burden using frame-wise expression labels derived from driving videos in the motion-transfer step, whereas most human facial expression datasets, such as VoxCeleb~\cite{Nagraniy2017}, require thorough manual annotation. The other annotation information, such as species (Japanese or Rhesus macaque), sex, and age, was provided as individual information in the dataset, while the annotations of head orientation were automatically estimated by InsightFace. For image editing based on age information, macaques under four years old were set as the negative group and the others were set as the positive group to determine the classification boundary and its normal vector.

\subsection{Exploration of disentangled latent representation in StyleSpace}\label{subsec:method-stylespace}

The ability of StyleGAN2 to disentangle visual features has the potential to identify shared facial parts' movements in several facial expressions across different individuals. To test the disentanglement ability of the trained model, we explored the latent space, specifically StyleSpace, of StyleGAN2, where the latent representation of human faces has been reported to be highly disentangled, with individual channels in particular layers corresponding to specific features~\cite{Wu2021}. We explored the two predefined distinct facial actions, that is, mouth and eye opening/closing movements. To identify the channels of StyleSpace related to mouth opening/closing and eye closing/opening representation, we analyzed the differential vector between the latent codes of the target facial movement images and the neutral face images, 
denoted by $\delta^e_r=\delta_{\mathrm{movement}}^e-\delta_{\mathrm{neutral}}^e$, where $\delta_{\mathrm{movement}}^e$ and $\delta_{\mathrm{neutral}}^e$ are the style vectors normalized by the population mean (the distribution of the entire face images) as defined in~\cite{Wu2021} for the target expressions and neutral expression, respectively. 
Calculating the differential vector relative to neutral face images enabled better editing axis extraction for the target facial movement than that without using differentiation. 
Moreover, to select the channels in StyleSpace that were well disentangled and devoted to a single movement, 
we manually categorized the 14 expression types from the training dataset into positive and negative samples depending on the presence and absence of particular facial movements, 
such as mouth opening/closing or eye closing/opening (the positive and negative samples of the two movements are shown in \tabref{posneg_def}). 
We then computed the mean measure of activity, $\theta_r$,
as follows and identified the channels with the highest value of $\theta_r$:

\eq{
  \theta_r = \frac{1}{N_p} \sum_{m_p} \theta_u^{(m)} - \frac{1}{N_n} \sum_{m_n}\theta_u^{(m)} ,
}

where $N_p$ and $N_n$ are the numbers of the expression types in the positive and negative samples, respectively, 
and $\theta_u^{(m)}$ denotes the relevance of channel $u$, 
as defined in the StyleSpace analysis~\cite{Wu2021} for one of 14 facial expression types.
In this study, we calculated $\theta_u$ using $\delta^e_r$ as $\theta_u=|\mu_u^e|/\sigma_u^e$, where $\mu_u^e$ and $\sigma_u^e$ are the mean and standard deviation of $\delta^e_r$, respectively.
We used a PyTorch implementation of the StyleSpace analysis~\footnote{\url{https://github.com/xrenaa/StyleSpace-pytorch}}, 
which has a better connection with the ReStyle framework than the official TensorFlow implementation. 
After calculating the relevance of all channels, 
we selected the top 5 channels for each movement and averaged their values across the test images 
to visualize how these channels selectively contribute to the representation of specific facial movements.

\section{Results}

\begin{figure*}[t]
  \centering
  \includegraphics[clip,width=0.8\linewidth]{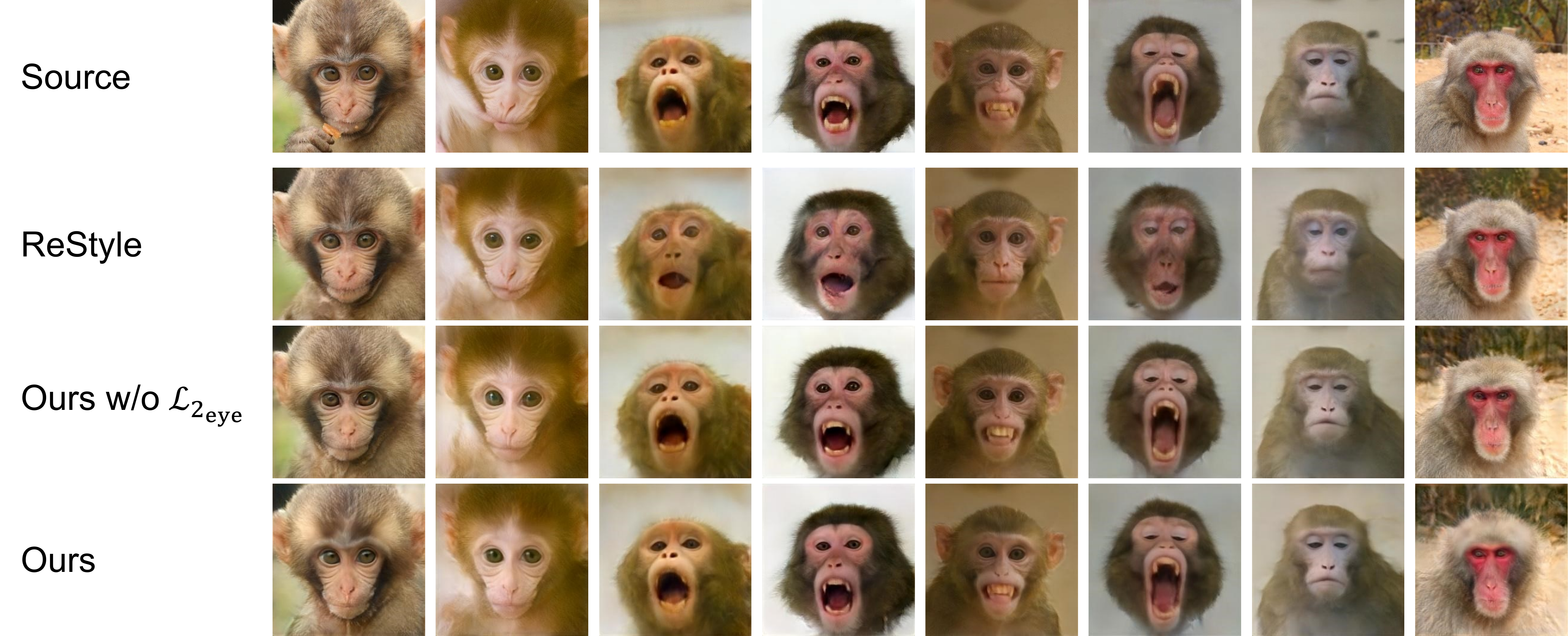}
  \caption{Qualitative comparison of the image inversion results using different training conditions. ReStyle~\cite{alaluf2021restyle} trained solely on the still image dataset failed to reconstruct facial expressions, particularly for images displaying closed eyes or open mouths, such as images in the sixth and seventh columns. By contrast, the proposed method successfully inverted the mouth movements and images showing exposed teeth, as indicated in the third and bottom rows. Moreover, the L2 loss for the eyes improved inversion quality around the eye regions including the reconstruction of eye movements as seen in the first column of the bottom row, which replicates the small sclera in macaques and depicts the eye looking to the left.}
  \label{fig:result-inv}
\end{figure*}

\subsection{Evaluation of the macaque face image generation model trained using the proposed method}
\subsubsection{Inversion quality}\label{subsec:result-inversion}

The inversion quality of the trained StyleGAN2 model on test images is an important measure of the model's ability to accurately represent input images, and is critical for the performance of downstream image editing tasks. The inversion results of representative test macaque images using several training procedures are shown in \figref{result-inv} for qualitative evaluation. The StyleGAN2 model trained solely on still macaque images without motion transfer-based data augmentation achieved poor inversion quality, particularly for images displaying closed eyes or open mouths, such as images in the sixth and seventh columns in \figref{result-inv}, 
which were outnumbered in the dataset. 
By contrast, our method with or without L2 loss for the eye region, successfully inverted images showing open mouth and exposed teeth, as indicated in the third and bottom rows in \figref{result-inv}. These results demonstrate that training with motion transfer-based data augmentation was effective for improving the inversion quality of face images with rich expressions. Moreover, the proposed method with L2 loss for the eyes accurately reproduced the eye movements, as seen in the first column of the bottom row, which replicates the small sclera in macaques and depicts the eye looking to the left.

\tabref{result-quantmotrans} presents the quantitative evaluation of the inversion quality, measured using the mean squared error (MSE) loss, LPIPS loss, and similarity loss (denoted by ID\_moco in \tabref{result-quantmotrans}) between the pixel values of the input and inverted images. We also included the MSE defined by different regions for this validation, along with the loss values used in the training procedure.

The LPIPS and ID\_moco losses from our training methods outperformed those from training solely on the still image dataset. This is likely to be because the motion transfer-based data augmentation balanced the facial expression distribution in the training dataset. For the MSE loss, the metric did not necessarily evaluate the facial expression reconstruction quality because it may reflect overall color degradation across the entire image. The MSE loss for the eye region still deviates from that outside the eye region, which indicates that the overall L2 loss weight tuning was insufficient to recover eye movements.

\tabref{result-quanteye} shows the effects of different weight values of the L2 loss for the eyes on the metrics while keeping the weight of the overall L2 loss at the default value of 1.0. 
The decrease in the MSE for the eyes indicates that our L2 loss for the eyes improved eye movements successfully in contrast to the L2 loss tuning alone in~\tabref{result-quantmotrans}.
Optimizing the weight of L2 loss for the eyes relative to the overall L2 loss led to a reduction in total MSE, although the MSE for regions outside the eyes increased slightly. Given that the inversion quality improved qualitatively when both total MSE and eye-region MSE were optimized, as shown in \figref{result-inv}, the poorer LPIPS and similarity loss values, which were based on models trained on object images other than macaque faces, did not appear to significantly affect inversion quality for macaque images.

\begin{table*}[t]
  \centering
  \caption{Quantitative evaluation of the impact of data augmentation via motion transfer on inversion quality, tested using 1,000 images. The first row indicates the final values of several error metrics for the model trained within the ReStyle framework only using the still image dataset with 120,000 training iterations. The second and third rows indicate the results of the model trained using our method using different weights of L2 loss with 300,000 iterations. The least losses in the columns are written in bold. We fixed $\lambda_{l2_{eye}}$ at the value of 0.0.}
  \label{tab:result-quantmotrans}
  \begin{tabular}{lccccc}
    \toprule
    & MSE$\downarrow$ & MSE$\downarrow$ (eye) & MSE$\downarrow$ (out of eye) & LPIPS$\downarrow$ & ID\_moco$\downarrow$ \\
    \midrule
   StyleGAN2 ($\lambda_{l2}=1.0$) & 0.0121 & 0.1390 & \textbf{0.0129} & 0.1571 & 0.0413 \\
   Motion transfer $+$ StyleGAN2 ($\lambda_{l2}=1.0$) & 0.0129 & 0.1137 & 0.0145 & \textbf{0.1447} & \textbf{0.0296} \\
   Motion transfer $+$ StyleGAN2 ($\lambda_{l2}=10.0$) & \textbf{0.0120} & \textbf{0.0998} & 0.0136 & 0.1473 & 0.0321 \\
   \bottomrule
  \end{tabular}
\end{table*}
\begin{table*}[t]
  \centering
  \caption{Parameter exploration results based on inversion quality tested using 1,000 images. Errors of the model trained using our method while varying the weights of the L2 loss for the eyes and fixing the weights of the overall L2 loss as 1, with 300,000 iterations are shown. The least losses in the columns are written in bold. We set $\lambda_{l2}$ at the value of 1.0.}
  \label{tab:result-quanteye}
  \begin{tabular}{lccccc}
    \toprule
    & MSE$\downarrow$ & MSE$\downarrow$ (eye) & MSE$\downarrow$ (out of eye) & LPIPS$\downarrow$ & ID\_moco$\downarrow$ \\
    \midrule
    Motion transfer$+$StyleGAN2 ($\lambda_{l2_{eye}}=5.0$) & 0.0222 & 0.0616 & \textbf{0.0166} & \textbf{0.1616} & \textbf{0.0323} \\
    Motion transfer$+$StyleGAN2 ($\lambda_{l2_{eye}}=10.0$) & \textbf{0.0203} & \textbf{0.0440} & 0.0169 & 0.1794 & 0.0336 \\
    Motion transfer$+$StyleGAN2 ($\lambda_{l2_{eye}}=50.0$) & 0.0281 & 0.0473 & 0.0253 & 0.2444 & 0.0446 \\
    \bottomrule
  \end{tabular}
\end{table*}

\begin{figure}[t]
  \centering
  \includegraphics[clip,width=0.7\linewidth]{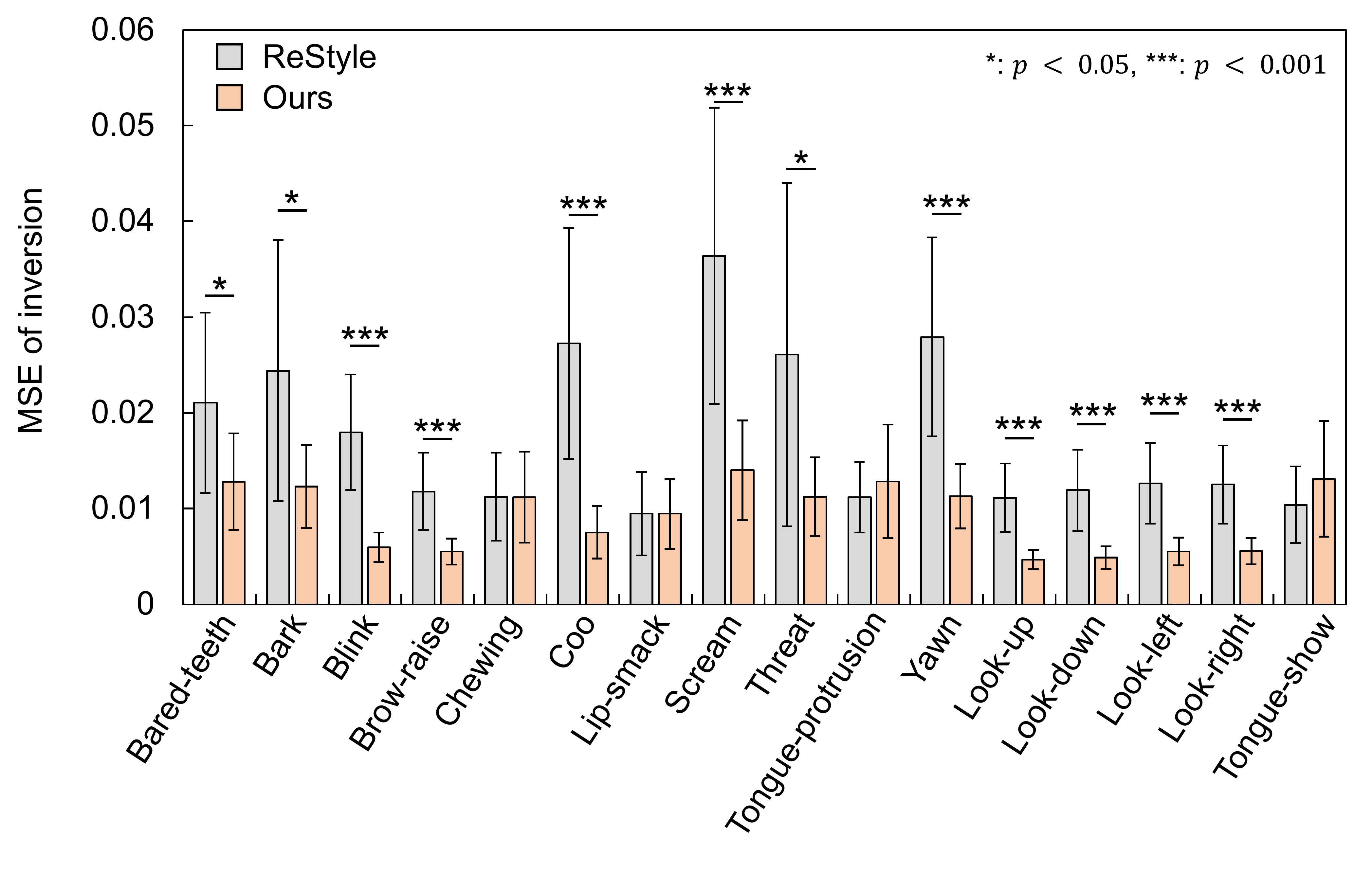}
  \caption{Inversion error for each facial expression. Errors were computed over 20 randomly selected test images inverted using the model trained only on still images with 120,000 iterations, labeled ``ReStyle'' and the model trained using our method for 320,000 iterations, labeled ``Ours''. Asterisks indicate the errors for each facial expression that showed a statistically significant difference between the two models, based on a pairwise \textit{t}-test with Holm adjustment.}
  \label{fig:result-cateeval}
\end{figure}

\figref{result-cateeval} shows the inversion quality based on MSE for individual facial expression types, 
comparing the standard ReStyle training condition 
using the still image dataset with the proposed method with the L2 loss for the eyes. 
MSE was calculated from images masked for the facial regions using \pleasecheck{a predefined mask as shown in Fig.~A2 in the supplementary material} to focus on evaluating the quality of the replicating facial movements rather than overall image fidelity including the region around the face area. The model trained using only still images failed to reconstruct facial movements related to mouth motions such as \baredteeth{}, \bark{}, \scream{}, and \threat{}, eye movements such as \blink{}, \browraise{}, \lookup{}, \lookdown{}, \lookleft{}, and \lookright{}, and jaw movements such as \coo{} and \yawn{}. 
By contrast, the model trained using the proposed method resulted in fewer errors for most facial expression types, 
which demonstrates the advantages of our approach. However, the inversion quality for \tongueprotrude{} 
and \tongueshow{} remained poor, even using our approach, 
probably because of poor motion transfer quality \pleasecheck{caused by a lack of training video data for these specific expressions, as shown in Fig.~A3.}

\begin{figure*}[t]
  \centering
  \includegraphics[clip,width=1.0\linewidth]{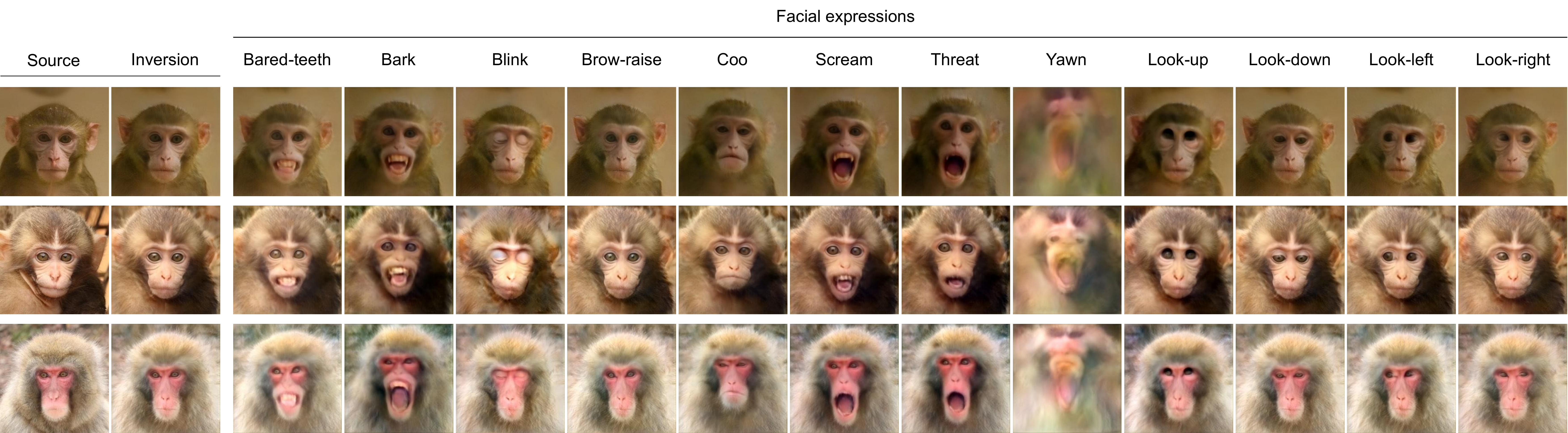}
  \caption{Editing results using annotation information about facial expression types. The editing strength was manually adjusted for each image and condition to produce the results shown in this figure.}
  \label{fig:result-edit-a}
\end{figure*}
\begin{figure}[t]
  \centering
  \includegraphics[clip,width=0.6\linewidth]{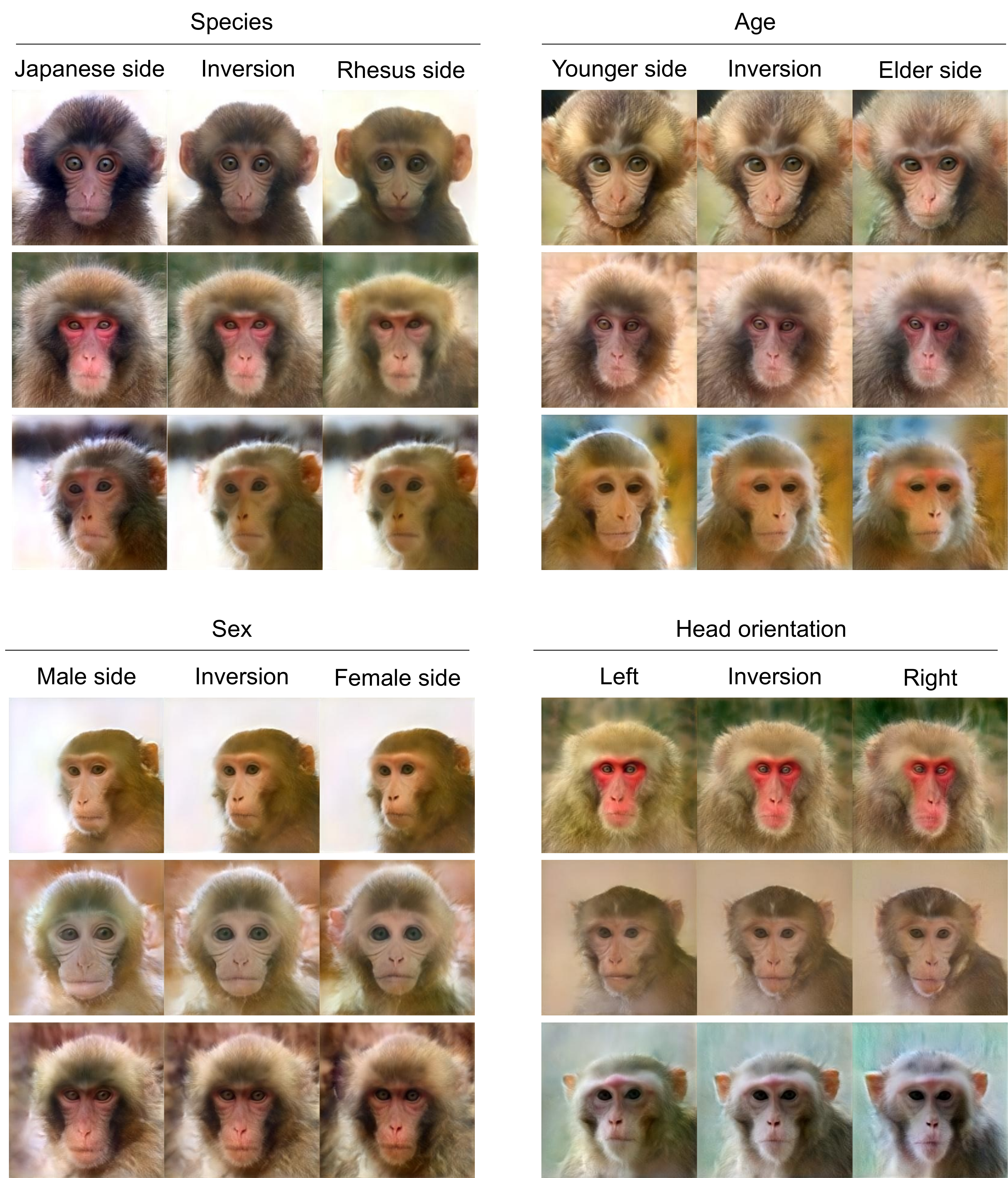}
  \caption{Editing results using annotation information from individual information.The editing strength was manually adjusted for each image and condition to produce the results shown in this figure.}
  \label{fig:result-edit-b}
\end{figure}

\subsubsection{Style-based image editing using annotation information}
If certain macaque facial attributes are represented linearly in the latent space of the trained StyleGAN2 model, then applying linear shift to the latent representation of a source image in the direction of the latent representation of images with a specific annotated attribute can serve as a method for editing the source image to enhance the annotated attributes. 
Because we used motion transfer-based data augmentation, the synthetic images generated through motion transfer retained labels related to facial expression attributes. Furthermore, some still images included individual details, such as identification, species (Japanese or Rhesus macaque), sex, and/or age. Head orientation was also estimated using InsightFace. These individual details from the still images were also retained in the motion-transfer synthetic images and made available as annotation information for style-based image editing
\begin{figure*}[t]
  \centering
  \subfloat[][Scatter plot illustrates the distribution of test images from 20 individuals across 14 expressions within the latent representation, defined by the mouth opening/closing axis ($x$-axis) and the eye closing/opening axis ($y$-axis).]
    {\includegraphics[clip,height=16.0\baselineskip]{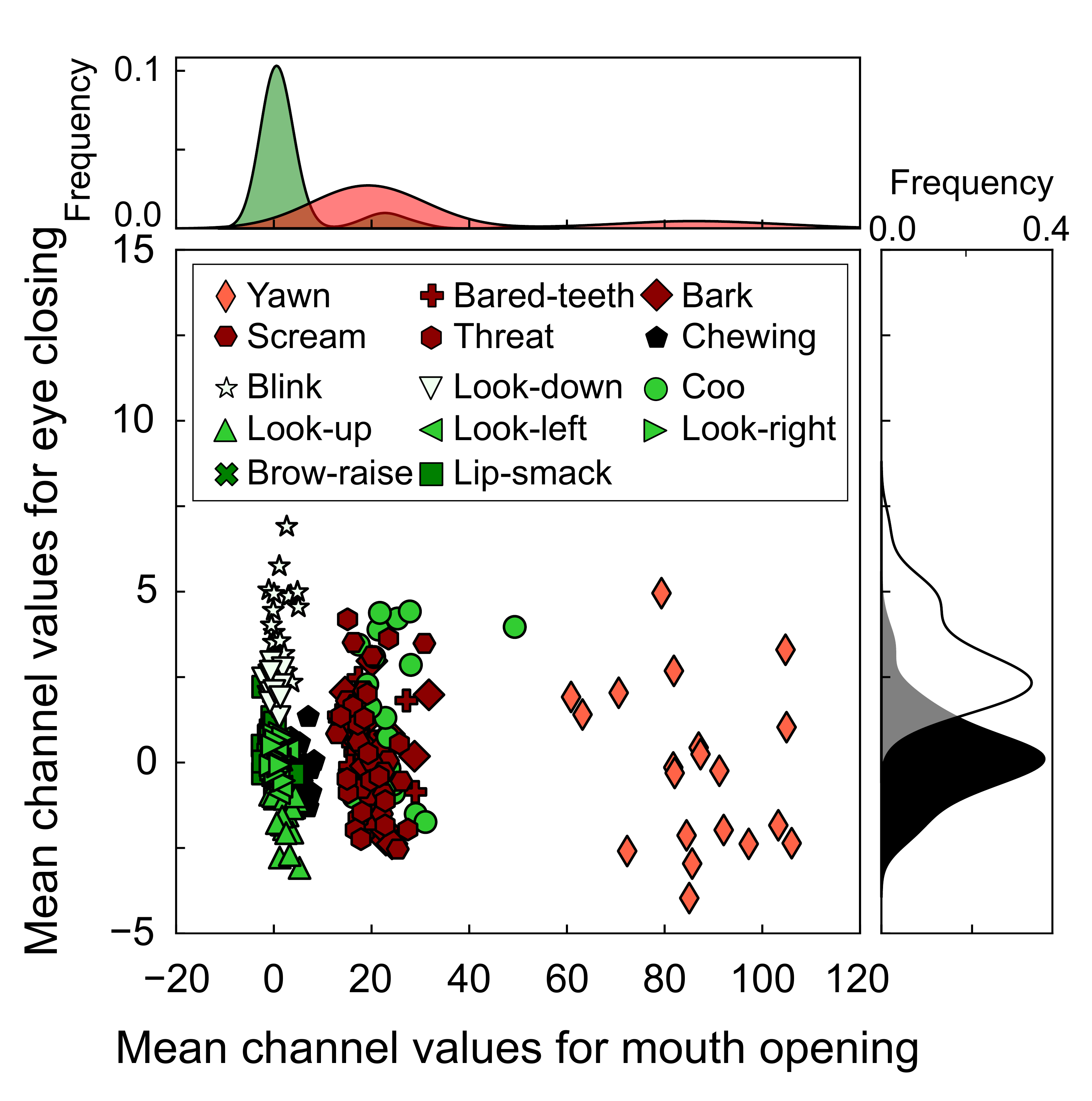}}\hspace{10mm}
  \subfloat[][Editing results of a test image are shown by modifying their latent codes along the mouth opening/closing axis and eye closing/opening axis.]
    {\includegraphics[clip,height=16.0\baselineskip]{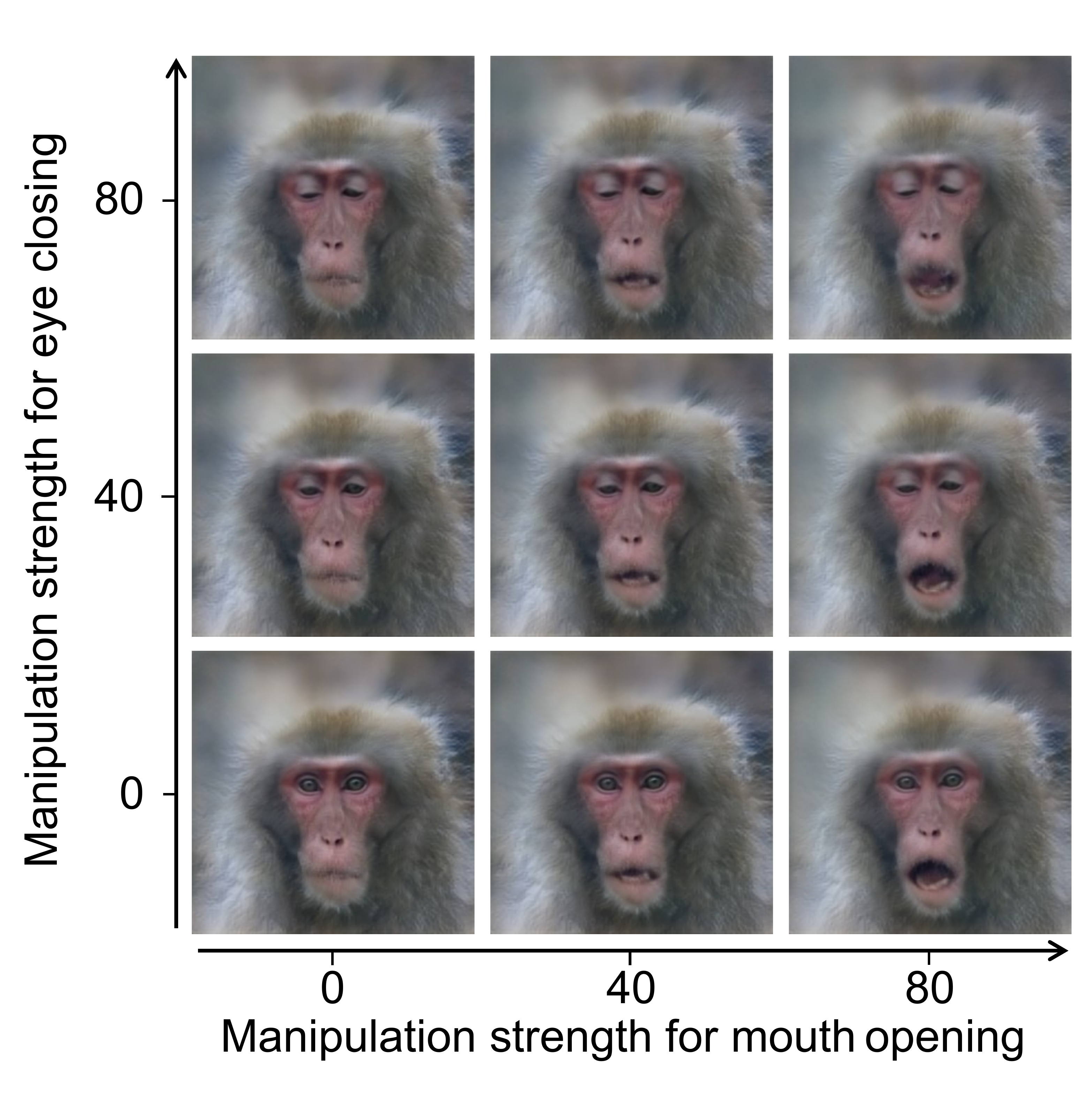}}
  \caption{Visualization of the mean values of the top 5 channels related to the mouth opening--closing axis and eye closing--opening axis. The output of the top 5 channels is commonly modulated by the group of facial expression types in the StyleSpace regardless of individual differences. In the scatter plot, greenish and reddish markers represent mouth closing and opening, respectively, whereas dark and pale markers represent eye opening and closing. The histograms at the top of the scatter plots represent the distribution of mean channel values for mouth-closing expressions (green) and mouth-opening expressions (red). The histograms on the right side of the scatter plots represent the distribution of mean channel values for eye-opening (black) and eye-closing (white) expressions.}
  \label{fig:result-chan}
\end{figure*}

\figref{result-edit-a} and \figref{result-edit-b} presents the editing results based on several attribute labels using the InterFaceGAN~\cite{Shen2022} procedure. 
\figref{result-edit-a} illustrates the editing results using facial expression labels and 
demonstrates good editing quality for a wide range of facial expression categories. 
For example, the \scream{} expression, which requires significant image changes to open the mouth, 
was reasonably replicated by simply adding a latent vector calculated from the images related 
to the \scream{} label to source images with a certain weight. 
Additionally, more subtle expressions, 
such as \blink{} and \lookleft{}, were successfully manipulated using the same editing procedure. 
However, some facial expressions, such as \yawn{}, 
failed to be manipulated using this method, which is likely to be because of an inconsistency in the latent representation of 
the \yawn{} expression, which made it difficult to estimate a reliable editing vector. 
\pleasecheck{Given the challenges in motion transfer (as shown in Fig.~A3 in the supplementary material),} 
the editing results demonstrate that the model trained using our method achieved 
a well-structured latent representation that can be edited linearly using attribute labels.

\figref{result-edit-b} shows the editing results using latent vectors between opposing labels in attributes related to individual information, 
such as species, age, sex, and head orientation. Editing results based on species information (Japanese--Rhesus axis) modified the face's contour, 
face size, and the skin and hair color of the source image to more closely align with the characteristics of the intended species. 
For example, edits directed toward Japanese macaques produced images with more reddish skin, gray hair, and an elongated face, 
whereas edits toward Rhesus macaques resulted in paler skin, yellowish hair, and a smaller face. 
Edits based on age information 
(younger--older axis) primarily changed the skin and hair color, 
in addition to the eye-to-face ratio, thereby reflecting typical age-related facial characteristics: 
older macaques exhibit more reddish skin, whiter hair, and a smaller eye-to-face ratio, whereas younger macaques have the opposite traits. 
For sex edits (male--female axis), changes included lower face height, jaw width, and nose length, 
with male macaques typically having larger features in these areas than females~\cite{Rosenfield2019}. 
Finally, edits based on head orientation information (left--right axis) effectively altered the head's direction 
in the source image toward the intended direction. 

\subsection{Extraction of editing axes in StyleSpace corresponding to motion components}

To evaluate the quality of disentanglement in the latent representation of the trained StyleGAN2 model and explore its potential for analyzing facial motion components, we investigated individual channels in the StyleSpace representation associated with the movements of specific facial parts, such as mouth opening--closing and eye closing--opening. We extracted the top 5 channels that were most modulated by specific facial part movements co-occurring across different types of facial expressions, regardless of individual macaque differences, using the method described in Section~\ref{subsec:method-stylespace}. 

\figref{result-chan} visualizes the mean value of the top 5 channels related to the mouth opening--closing axis and 
eye closing--opening axis in response to different facial expression images from different individuals. 
The scatter plot in \figref[a]{result-chan} shows that various facial expressions of different macaque individuals 
indeed are distributed along these two axes according to the intended facial part movements, 
although each axis is defined only by the output of five channels. 
\figref[b]{result-chan} illustrates that a systematic change in these two axes results in image generation with the intended eye and mouth movements, 
which indicates that the trained StyleGAN2 model successfully disentangled features related to specific facial parts into a few channels.
\pleasecheck{Moreover, the inspection of the inversion results after modifying individual channels' outputs, as shown in Fig.~A3 in the supplementary material, demonstrated that some channels were dedicated 
to controlling specific facial attributes}, such as opening the mouth, the shape of the top of the head, eye direction, and eye size. 

\section{Discussion}

In this study, we developed a motion transfer-based data augmentation technique to increase both the quantity and diversity of training images in a dataset, along with modifications to ensure a wide variation and improve the quality of image generation. With this new technique, we addressed the limitations of previous StyleGAN2 training for facial expressions of non-human primates: image databases with diverse facial expressions are scarce in comparison with human databases. We included two species of macaques as references, incorporating diverse individual features, such as identification, sex, age, and population origin, which ensured a well-represented source dataset. The data augmentation approach using motion transfer was effective because, although most of the still images used as source images featured neutral expressions, facial movements are highly transferable across different individuals due to their musculoskeletal similarity. This augmentation enhanced the inversion and editing quality of StyleGAN2 for a variety of facial expressions and individuals. To the best of our knowledge, using motion transfer techniques for data augmentation to improve the generation and editing quality of image generative models for animal facial expression is a novel approach. A key advantage of this method, alongside data augmentation, is that it allows us to incorporate annotation information about driving movements into the synthetic outputs, thereby aiding further editing and enabling face image analysis through supervised learning. Additionally, the method of evenly sampling images based on the tentatively acquired latent representation was useful for distilling informative images from real videos for training. To accurately reconstruct subtle but critical facial changes for social communication in primates, such as eye movements, which are more challenging to detect for macaques than for humans because of smaller or absent visible scleras, we designed loss functions that included a pixel loss specifically for the eye regions. The results demonstrated the ability of the StyleGAN2 model trained using our method to generate a diverse range of facial expressions, from subtle eye movements (e.g., \lookright{}) to more conspicuous facial displays (e.g., \yawn{}). 

We also found that the latent space of the trained StyleGAN2 model was linearly disentangled, to a degree, to enable facial images to be edited along a single axis corresponding to an annotated attribute, such as facial movements, species, sex, age, and head orientation. The linear editing of the latent representation can synthesize facial expressions onto images of new macaque individuals. The diversity of facial expressions and identities represented within the latent space of the trained model, and the possibility of generating diverse macaque face stimuli, have direct applications for behavioral and neuroscientific research. For example, by manipulating the latent space, we can generate a novel macaque face or a morphed face that combines two individuals \pleasecheck{(See Fig.~A1 in the supplementary material for an example of morphing results)} to explore the behavioral or neural responses to the perception of face identity, sex, age, or facial expressions. To better understand primate social communication and expression, experiments with naturalistic and ecologically valid face stimuli are essential, therefore this approach can contribute to advances in the facial expression field. 

To investigate the potential of our model for the analysis of movement in specific facial features, we explored the specificity of individual channels in the StyleSpace latent representation for eye and mouth opening/closing. These movements were distinctively represented by the outputs of only the top 5 channels in the latent space, which can be potentially used for the automatic detection of mouth and eye movements. 
Because the trained StyleGAN2 has successfully disentangled various facial components in its latent space, in the future, it may be possible to extract latent space axes corresponding to AUs. 
Additionally, using a dataset with annotations of AUs, even a small dataset, could suffice because the training data could be augmented using our proposed method (see the following Section~\ref{subsec:macaqueFACS} for details.).

\section{Limitations and Future Works}

\subsection{Limitations of the present study}

In this paper, we focused on Japanese and Rhesus macaques, two commonly used non-human primate species for translational research in animal behavior and neuroscience, which bridges the gap between human and non-human animals. However, we were unable to train the StyleGAN2 model on other primate species because of the lack of publicly or readily accessible video and still image datasets. We believe that the method developed in this study is applicable to other primate species beyond the two reference macaque species. This is because the underlying assumption for our method, that is, the similarity in musculoskeletal structure across individuals and the use of typical face expressions for communication within the same species, should be transferable to other primates. Furthermore, our cross-species trained StyleGAN2 model can generate highly naturalistic face images of two different species of macaques, despite some facial morphological differences. This aspect indicates the potential use of the trained model to identify facial features characteristic of either species by extracting channels that are selectively activated for one species or the other. A simple analysis of this application was described in Section~\ref{subsec:method-stylespace}, where one species was set as positive samples and the other as negative samples.

Some of the editing results in \figref{result-edit-a}, such as \yawn{}, 
were of poor quality because of insufficient motion transfer quality in certain samples or because of the nonlinearity of the latent representation of 
complex facial expressions, which prevented the accurate detection of the manipulation axis for the target expression using a simple editing method, such as InterFaceGAN. 
\pleasecheck{Additionally, the motion transfer model failed to transfer some facial expressions, such as \coo{} and \tongueprotrude{} as shown in Fig.~A3 in the supplementary material.} 
To improve the quality of both motion transfer and StyleGAN2, 
incorporating additional real video data that includes these missing facial movements for training would improve the model. 
Even if the additional videos mainly featured macaques with neutral expressions during naturalistic behavior, our resampling 
approach using a latent representation of a tentatively trained StyleGAN2, was crucial for expanding the video dataset to efficiently 
include less frequent facial expressions. 
In the present study, we used CG videos generated by MF3D as driving videos 
for facial expression-based data augmentation. 
Because CG videos do not perfectly capture naturalistic macaque facial expressions, 
the synthesized images produced through motion transfer also deviated from a natural appearance. 
As a result, the editing outcomes based on the annotation information for facial expression categories appeared unnatural in several respects. 
Using real videos, selected to represent a variety of facial expressions, 
as driving videos is likely to improve the naturalism of the synthesized training images and produce more realistic editing results. 
The application of a more refined editing method than that used in the present study may also be required if more naturalistic control 
over facial expressions is necessary.

\subsection{Potential application of the developed model}\label{subsec:macaqueFACS}
The advantage of using StyleGAN2 and its encoder for face image generation is that facial features are hierarchically disentangled in the latent space,
which could be used for automatic facial expression analysis. 
Animal FACS are anatomical tools used to study facial movements in animals~\footnote{\url{https://animalfacs.com/}} and were developed for various species by adapting human-based FACS, including domestic animals such as dogs and cats, in addition to various primates~\cite{caeiro2013orangfacs,correia2022callifacs,Vick2007}. Specifically for macaques, MaqFACS~\cite{correia2021extending,Parr2010} was developed based on anatomical knowledge and video images of Japanese and Rhesus macaques. With the aid of MaqFACS, human and macaque AUs have recently been compared numerically~\cite{kavanagh2022revisiting,Taubert2021}. This facial motion system is also useful for welfare applications, for example, estimating macaque emotions by predicting AUs from facial images~\cite{Morozov2021}. 
The automated analysis of macaque faces would free researchers from manually identifying AUs based on FACS in video clips, 
a task that is very time-consuming and requires training and certification for each species of interest~\cite{Ekman2002,Parr2010}. 
This would enable researchers to quantitatively evaluate facial expression changes over long periods across multiple 
individuals and in large samples and lead to improving the understanding of social communication and expression. 
Additionally, the well-organized representation of facial expressions in the latent space is potentially useful 
for estimating internal states of macaques using AUs~\cite{Morozov2021}, with practical applications in animal welfare. 
Combining the quantitative evaluation of behavioral cues, such as facial movements, with physiological changes, such as hormones~\cite{Correia-Caeiro2024} 
or heart rate variability~\cite{Katayama2016}, may allow us to estimate animal emotions, which are likely to differ  from human emotion expressions~\cite{Catia2017,Kret2020} 
and are not easily identified by humans~\cite{Correia-Caeiro2020,Marechal2017}. 
Comparing latent representations of macaque and human faces could also provide insights for translational research on facial expression and cognition. 
A promising approach for precise AU analysis using our method would be to create CG video clips of all AUs, 
incorporating anatomical knowledge of the muscular movement and bone structures, such as the 3D CG parameters provided by MF3D. 
The synthetic datasets using motion transfer driven by these CG video clips could then be used for training the StyleGAN2 model with annotated AUs. 
The correspondence between the disentangled facial features at the level of channels of style parameters and AUs annotations 
or the anatomical parameters would provide a more detailed disentangled latent representation of macaque faces, thereby enabling finer image editing control and more accurate facial analysis.
\section{Conclusion}

In this study, we presented a new framework for training a StyleGAN2 model to generate non-human primate face images, 
specifically for Japanese and Rhesus macaques, with rich variations in expression, even when the training datasets lack diversity. 
Our results demonstrated the capability to reconstruct images with various facial expressions across various identities 
and to edit images by manipulating the latent space, which was not achieved using previous methods. 
This suggests that the trained StyleGAN2 model could be applicable to the future automatic detection of AUs based on FACS, 
a crucial tool for animal behavior research. 
A highly descriptive image generation model is also valuable for creating diverse facial images, 
thereby enabling behavior and neuroscience researchers to explore the facial recognition process and reactions to these images.

\section{Author contributions}
\RH{} conceived and supervised the study. 
\TI{} and \RH{} designed the methods and experiments. 
\AY{} contributed to designing the method. 
\TI{} implemented the methods and conducted the experiments. 
\RH{}, \TMN{}, and \CCC{} provided data. 
\TI{} and \RH{} drafted the manuscript. 
All authors revised the manuscript. 

\section*{Acknowledgment}
\addcontentsline{toc}{section}{Acknowledgment}
This work was supported in part by the Japan Science and Technology Agency, Moonshot Research \& Development Program grant JPMJMS2012 and the National Institute of Information and Communications Technology (NICT) grant NICT 22301, and MEXT/JSPS KAKENHI Grant-in-Aid for Transformative Research Areas (A), Grant Number 24H02185 and Grant-in-Aid for Scientific Research (B), Grant Number 24K03241. 


\bibliographystyle{unsrtnat} 
\bibliography{reference}

\onecolumn

\section*{Appendix}

\subsection{Additional information on the training conditions}

\subsubsection{Training results of the thin-plate spline motion model (TPSMM)}

We conducted a quantitative evaluation of TPSMM using 258 test video clips that were reserved separately from the training data. 
The L1 loss results from the motion transfer for each expression are shown in \tabref{appendix-evaltpsmm}. 
Because there is no definitive landmark detector or identity evaluator network for macaques, we did not conduct evaluations with AKD, MKR (key point detection accuracy evaluation), or AED (identity identification accuracy evaluation)~\cite{Siarohin2021,Zhao2022}, and performed the evaluation only using the L1 loss. By changing the number of thin-plate splines to $K = 10, 20, 30, 40$, and $50$ during training, we found that the L1 loss for the test data decreased at $K = 30$ and $40$. 
Because these are the motion transfer results for a limited number of individual macaque images, 
we also qualitatively confirmed the results against the images selected from the \textit{10\_monkey\_species/n3} archive. As shown in \figref{appendix-motrans}, when $K = 30$ and $40$, the quality of motion transfer for new images was nearly at the same level; hence, to enhance key point detection performance, we set the parameter value to the smaller $K = 30$. Finally, to achieve better convergence, we increased the number of iterations from the default value of 75 to 150 to obtain the final training checkpoint.

As for the qualitative evaluation of the motion transfer results, we show representative results for transferred motion of various facial expressions in \figref{appendix-motrans}. 
The synthetic images generated by the TPSMM model are shown, with the number of iterations as 75 (default value) by changing the number of thin-plate splines to $K=10, 20, 30$, and $40$. 
The motion transfer quality improved prominently using $K$ larger than 20. 
When $K=40$, some images exhibited blurring compared with $K=30$. 
Blurring occurred with the \yawn{} expression, which was scarce in the training data, but involved large mouth movements.



\subsubsection{Image dataset for the first round of StyleGAN2 training}

To expand the number of training images, the first round included 6,607 images sourced from LAION-5B~\footnote{\url{https://laion.ai/blog/laion-5b/}} using a keyword search for ``macaque monkey,'' in addition to \pleasecheck{the still image dataset described in Section~3.2.1}. Training the model on a larger still image dataset would help to improve its image generation performance. However, we excluded these LAION-5B images from the second round because they were not precisely categorized into Japanese and Rhesus macaques, and were contaminated with images of other monkey species. The dataset for the first round comprised still images and their motion-transferred images, driven by all frames of 16 driving videos (total 686,747 frames). Although this dataset was mostly biased toward faces with neutral expressions, the StyleGAN2 model trained in the first round was useful for selecting macaque face images based on similarity in latent space, which we then used for the second round of training.

\subsubsection{Masked regions used for validating the inversion quality of individual expression types}

\pleasecheck{Mask images for evaluating the inversion quality used in Fig.~4} 
and \tabref{appendix-error} are shown in \figref{appendix-mask}. Because the face alignment was performed before StyleGAN2 training, we can define mouth and eye areas of macaque faces using these mask images to calculate the error in the inversion of test images. 

\subsubsection{Example images for the second round of training the StyleGAN2}

\begin{figure*}[t]
  \centering
  \subfloat[][$K=10$.]
    {\includegraphics[clip,width=0.49\linewidth]{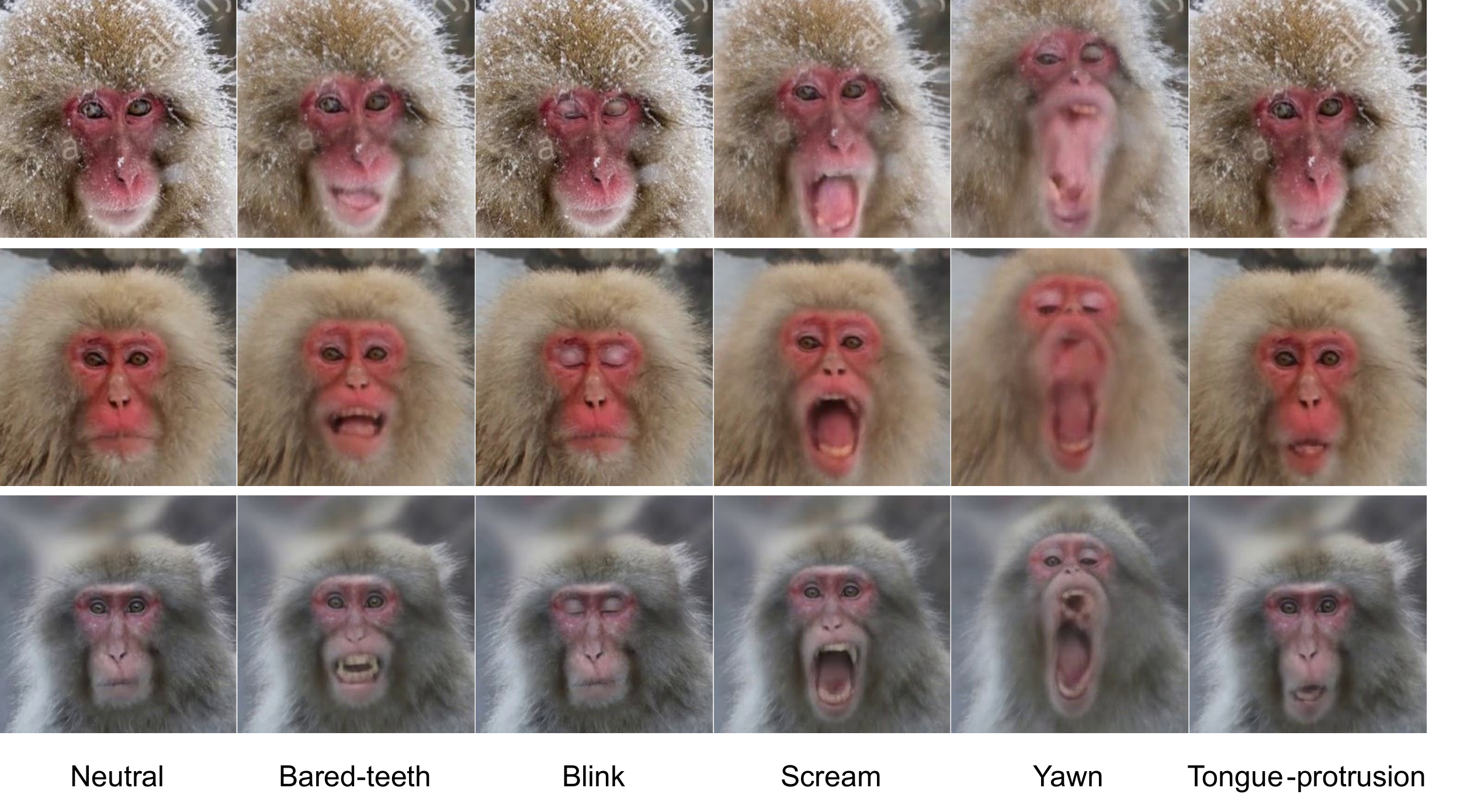}}
  \subfloat[][$K=20$.]
    {\includegraphics[clip,width=0.49\linewidth]{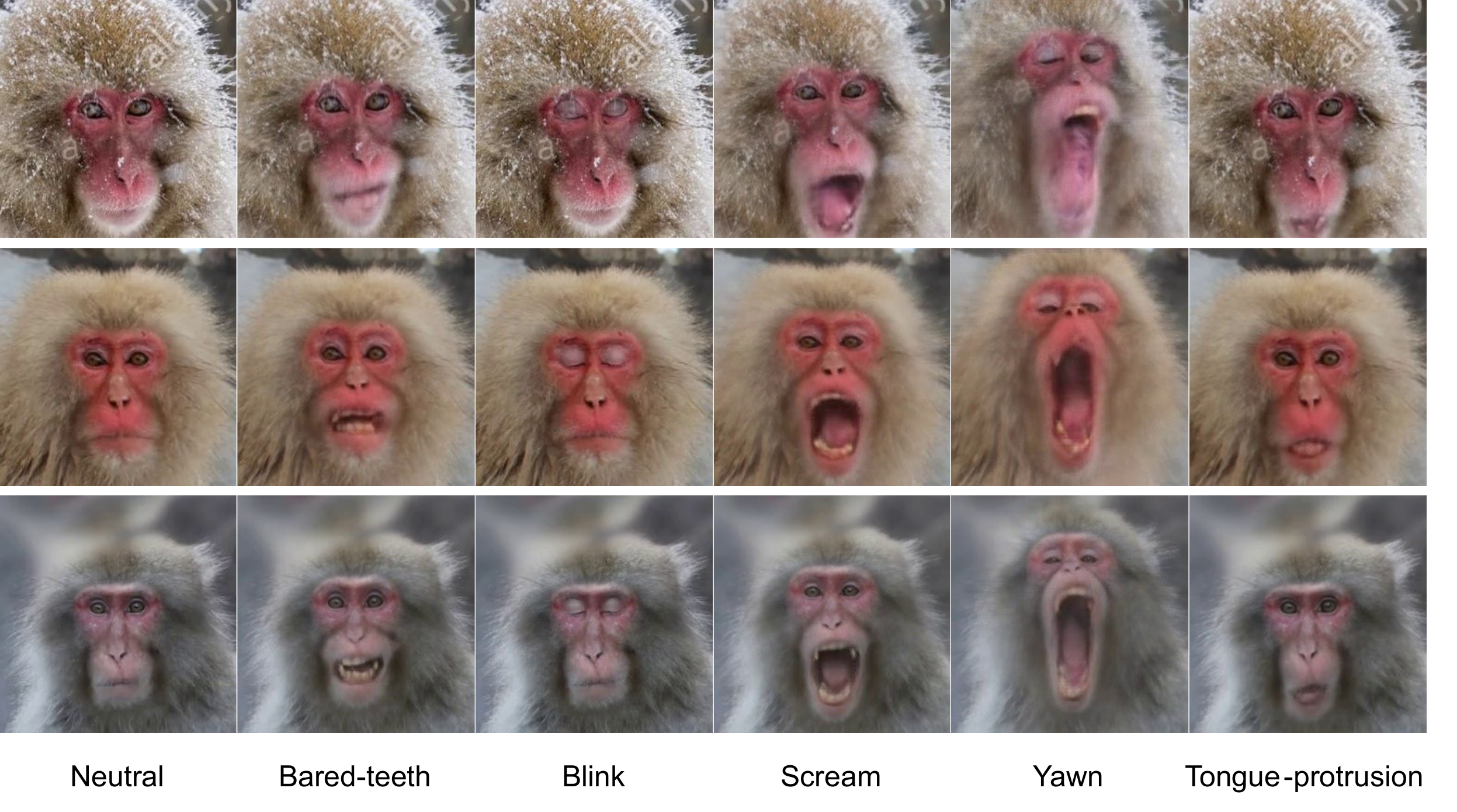}}\\
  \subfloat[][$K=30$.]
    {\includegraphics[clip,width=0.49\linewidth]{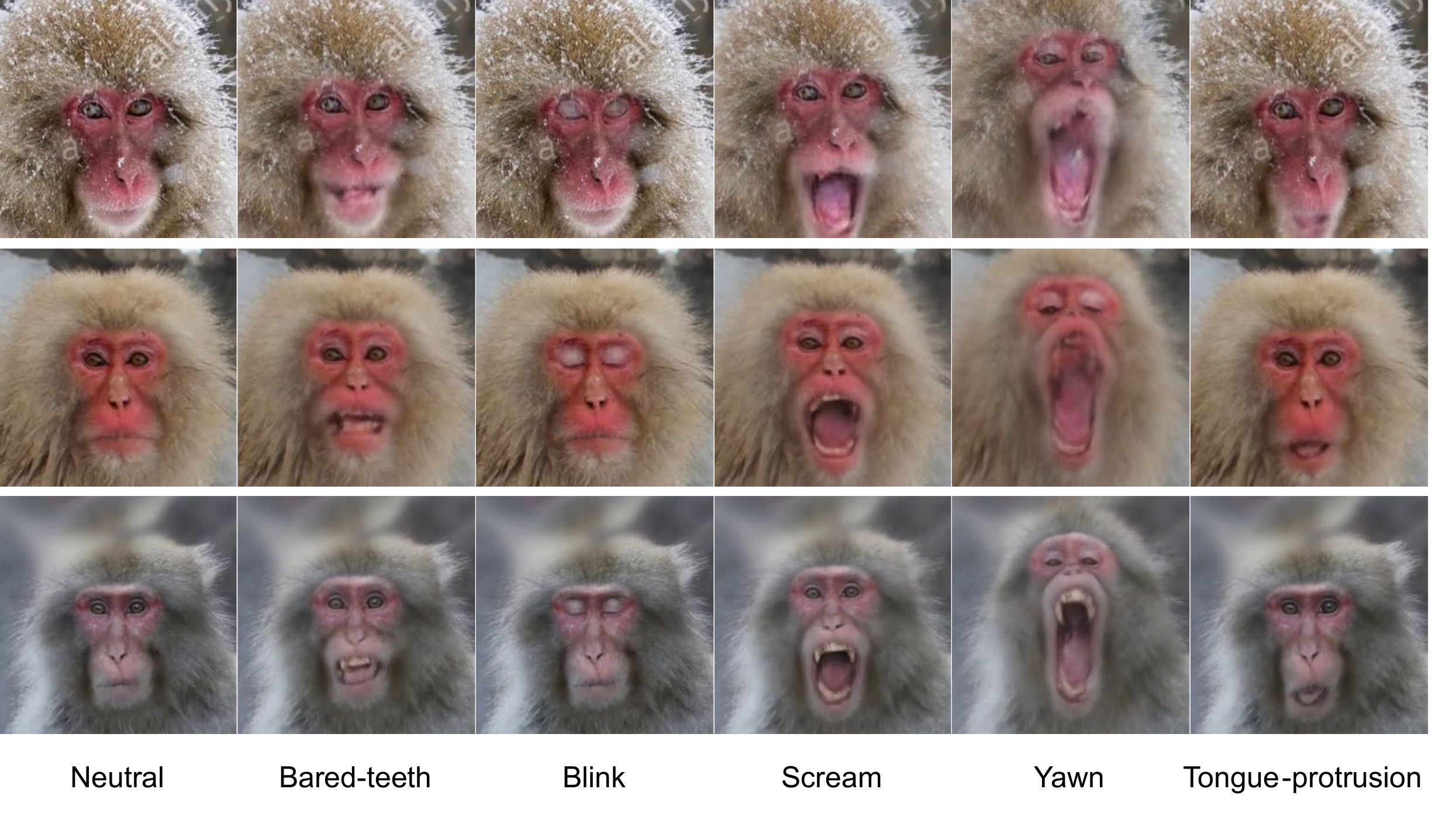}}
  \subfloat[][$K=40$.]
    {\includegraphics[clip,width=0.49\linewidth]{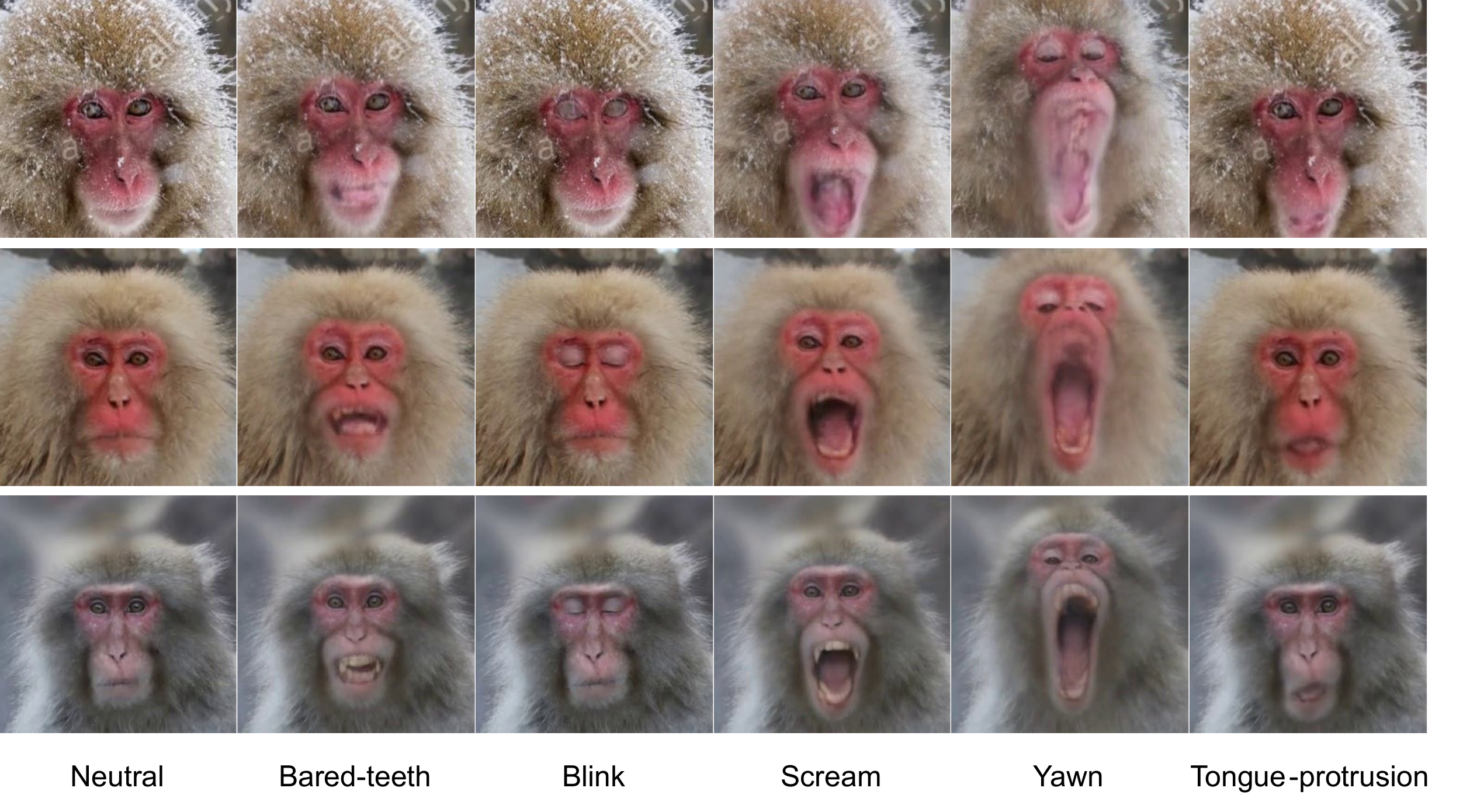}}
  \caption{Qualitative evaluation of the motion transfer results by varying the number of splines ($K$), while fixing the number of iterations at 75. The source still images were selected from the \textit{10\_monkey\_species/n3} archive. The synthetic images that correspond to the frame of the largest facial movements for each expression are shown in the figure. Each column corresponds to (from left to right) source image, \baredteeth{}, \blink{}, \scream{}, \yawn{}, \tongueprotrude{}.}
  \label{fig:appendix-motrans}
\end{figure*}

\begin{figure*}[t]
  \centering
  \subfloat[][A mask image for mouth-related expressions.]
    {\includegraphics[clip,height=10\baselineskip]{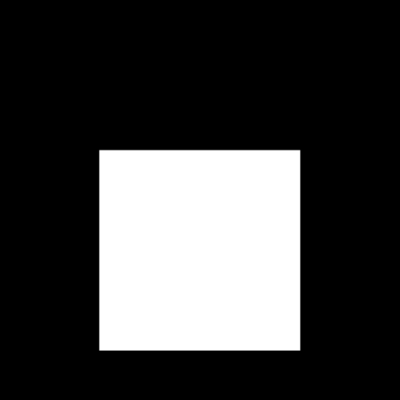}} \hspace{6mm}
  \subfloat[][A mask image for eye-related expressions.]
    {\includegraphics[clip,height=10\baselineskip]{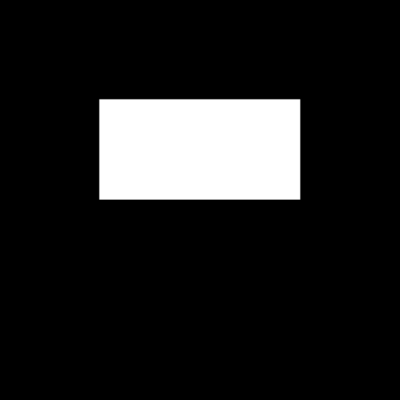}} \hspace{6mm}
  \subfloat[][A mask image for expressions involving entire face movements.]
    {\includegraphics[clip,height=10\baselineskip]{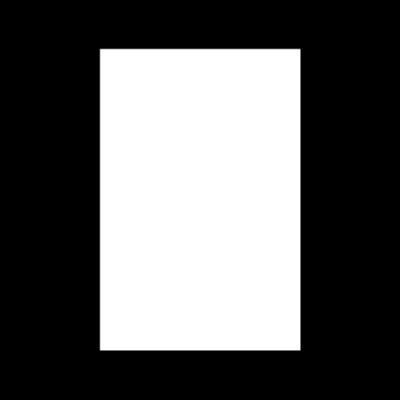}}
  \caption{Mask images applied to evaluate the inversion quality of individual expression types. a) A mask image for \baredteeth{}, \bark{}, \chewing{}, \lipsmack{}, \scream{}, \threat{}, \tongueprotrude{}, and \tongueshow{}, b) a mask image for \blink{}, \browraise{}, \lookup{}, \lookdown{}, \lookleft{}, and \lookright{}, c) a mask image for \coo{} and \yawn{}.}
  \label{fig:appendix-mask}
\end{figure*}

\begin{table}[t]
  \centering
  \caption{Mean L1 pixel losses over the frames in 258 test video clips using the TPSMM trained with 4,934 video clips. The evaluation was conducted while varying the number of splines ($K$) and the number of iterations. Note that $K=10$ was used for training TPSMM on human face images from the VoxCeleb dataset~\cite{Nagraniy2017}.}
  \label{tab:appendix-evaltpsmm}
  \begin{tabular}{cccc}
    \toprule
    & & \multicolumn{2}{c}{Image repeats per epoch} \\
    \midrule
    & & 75 & 150 \\
    \multirow{5}{*}{\shortstack[l]{Number of \\thin-plate spline motion ($K$)}} & 10 & 0.01676 & - \\
    & 20 & 0.01650 & - \\
    & 30 & 0.01613 & 0.01582 \\
    & 40 & 0.01612 & - \\
    & 50 & 0.01607 & - \\
    \bottomrule
  \end{tabular}
\end{table}

\begin{figure*}[!ht]
  \centering
  \subfloat[][ Example images randomly sampled from still image datasets. The StyleGAN model trained using only these still images served as the baseline to evaluate our proposed method.]
    {\includegraphics[clip,width=1.0\linewidth]{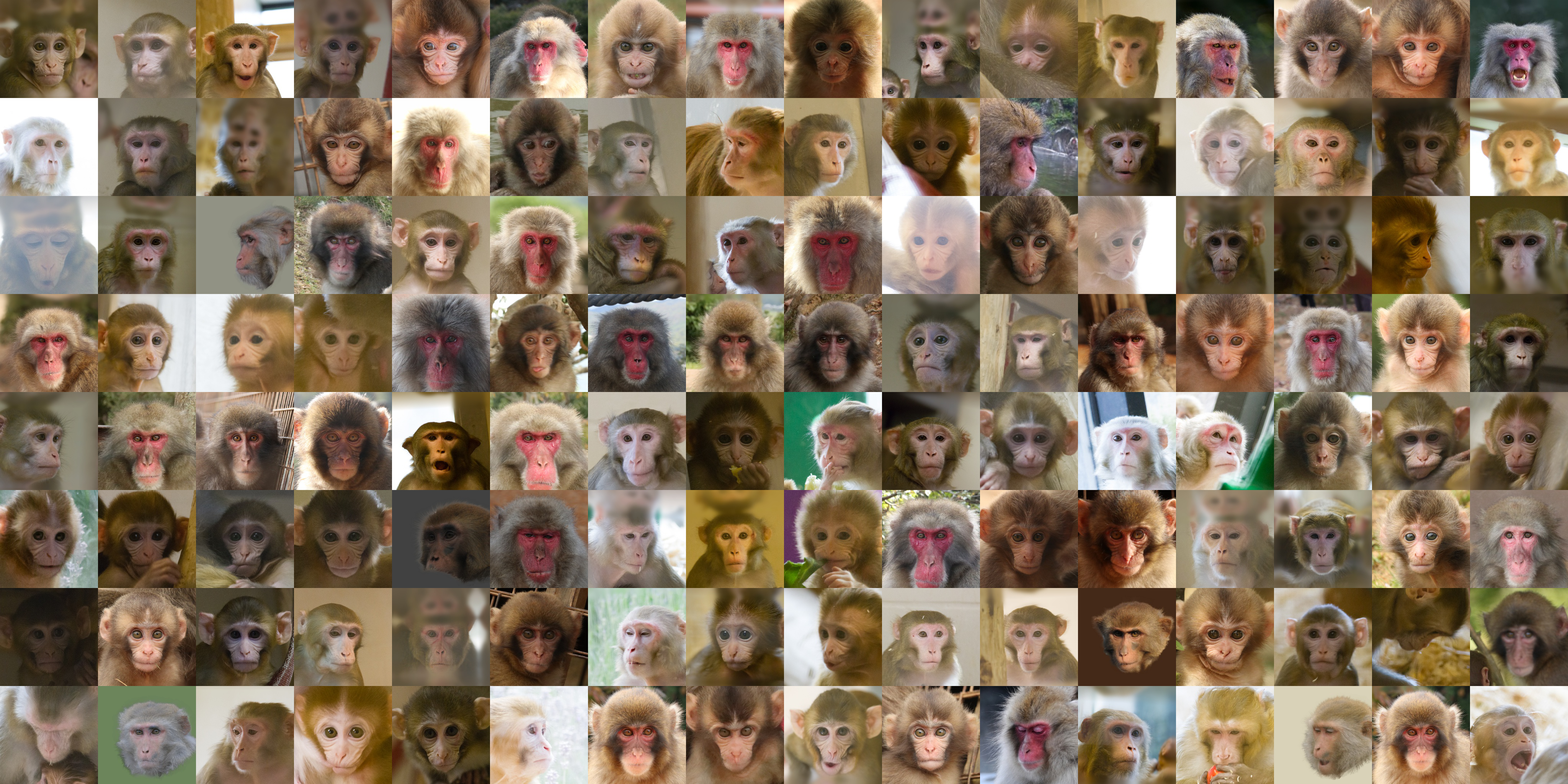}}\\
  \subfloat[][Example images randomly sampled from those used in the second round of training StyleGAN2 using our proposed method.]
    {\includegraphics[clip,width=1.0\linewidth]{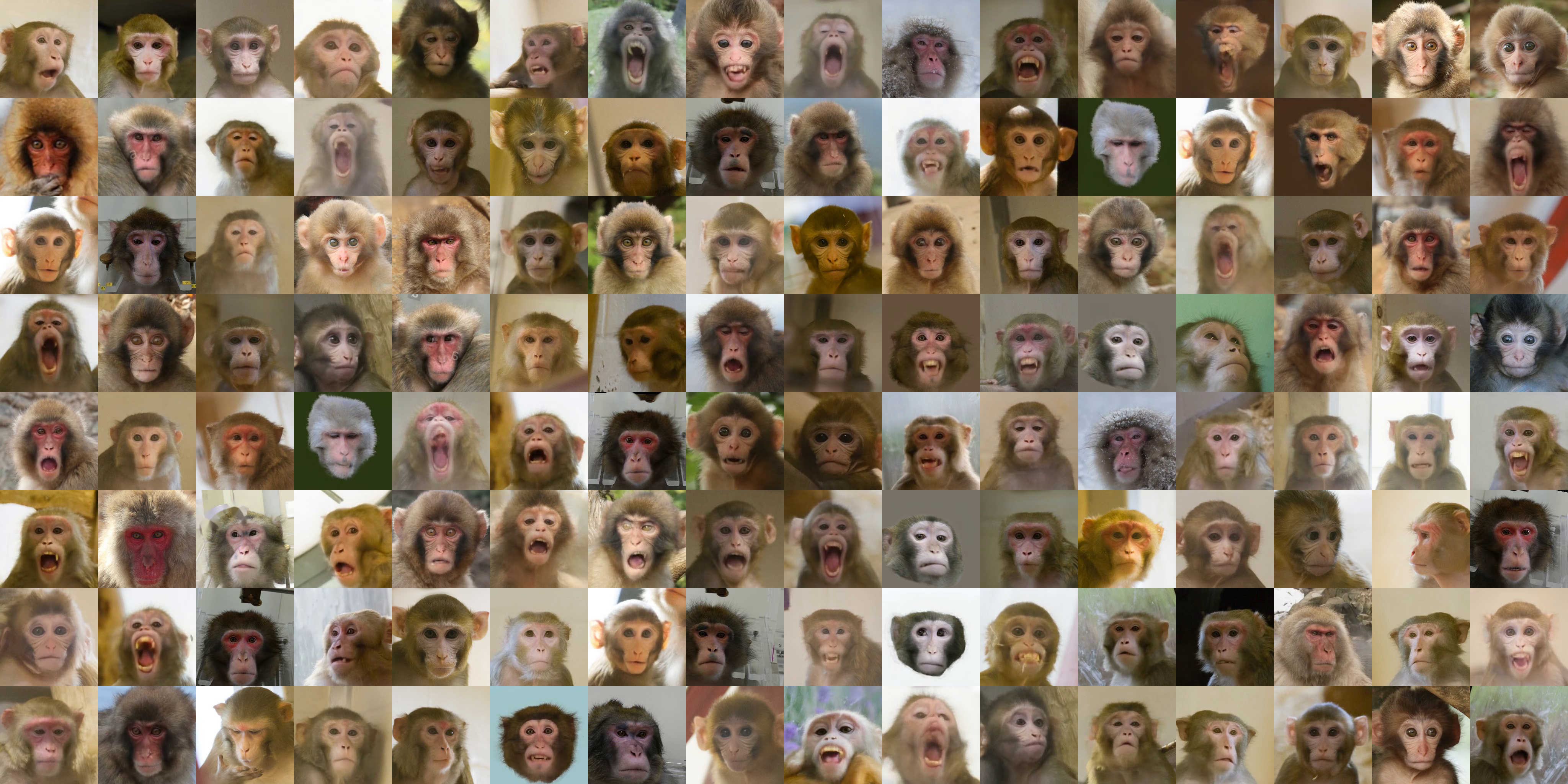}}
  \caption{Images used for the training of StyleGAN2 within the ReStyle framework.}
  \label{fig:appendix-example}
\end{figure*}
The variation of facial expression was augmented and the bias toward neutral faces in the training dataset was mitigaed by the proposed method. We show 480 randomly selected examples from the image dataset for StyleGAN2 training in \figref{appendix-example}. The proposed method enabled model training using the dataset that contained eye and mouth movements more frequently than using the original still image dataset.

\newpage
\newpage\phantom{blabla}
\newpage\phantom{blabla}
\newpage 
\begin{table*}[t]
  \centering
  \caption{Inversion error (MSE) for each facial expression. For comparison, a model trained using only the still image dataset with the ReStyle framework was used as baseline control. The images annotated with facial expressions for this comparison were synthesized images using the transfer model trained on our video dataset. Errors were computed based on 20 randomly selected test images. In our proposed method, the weight of the loss for the eyes was set to 0 for the condition without eye loss and 10.0 for the condition with eye loss.}
  \label{tab:appendix-error}
  \begin{tabular}{lccc}
   \toprule
   Facial expressions & ReStyle & Ours w/o $\mathcal{L}_{{l2}_{eye}}$ & Ours \\
   \midrule 
   \baredteeth{} & 0.02105 $\pm$ 0.00943 & 0.01418 $\pm$ 0.00509 & 0.0128 $\pm$ 0.00504 \\
   \bark{}       & 0.0244 $\pm$ 0.01366 & 0.01331 $\pm$ 0.00494 & 0.0123 $\pm$ 0.00433 \\
   \blink{}      & 0.01798 $\pm$ 0.00603 & 0.02087 $\pm$ 0.00689 & 0.00597 $\pm$ 0.00154 \\
   \browraise{}  & 0.0118 $\pm$ 0.00402 & 0.01557 $\pm$ 0.00539 & 0.00551 $\pm$ 0.00138 \\
   \chewing{}    & 0.01125 $\pm$ 0.00461 & 0.01287 $\pm$ 0.00451 & 0.01121 $\pm$ 0.00475 \\
   \coo{}        & 0.02726 $\pm$ 0.01205 & 0.01046 $\pm$ 0.00378 & 0.00753 $\pm$ 0.00273 \\
   \lipsmack{}   & 0.00947 $\pm$ 0.00434 & 0.01137 $\pm$ 0.00443 & 0.00946 $\pm$ 0.00366 \\
   \scream{}     & 0.03639 $\pm$ 0.01547 & 0.015 $\pm$ 0.00536 & 0.014 $\pm$ 0.00519 \\
   \threat{}     & 0.02607 $\pm$ 0.01791 & 0.01247 $\pm$ 0.00468 & 0.01124 $\pm$ 0.00412 \\
   \tongueprotrude{} & 0.01119 $\pm$ 0.00366 & 0.015 $\pm$ 0.00524 & 0.01284 $\pm$ 0.00594 \\
   \yawn{}       & 0.02792 $\pm$ 0.0104 & 0.014 $\pm$ 0.00441 & 0.0113 $\pm$ 0.00334 \\
   \lookup{}     & 0.01115 $\pm$ 0.00359 & 0.01396 $\pm$ 0.00407 & 0.00467 $\pm$ 0.001 \\
   \lookdown{}   & 0.01191 $\pm$ 0.00424 & 0.01496 $\pm$ 0.00517 & 0.00487 $\pm$ 0.00118 \\
   \lookleft{}   & 0.01263 $\pm$ 0.00421 & 0.01602 $\pm$ 0.00481 & 0.00551 $\pm$ 0.00144 \\
   \lookright{}  & 0.01251 $\pm$ 0.00409 & 0.01598 $\pm$ 0.00521 & 0.00557 $\pm$ 0.00135 \\
   \tongueshow{} & 0.01041 $\pm$ 0.00401 & 0.01329 $\pm$ 0.005 & 0.01312 $\pm$ 0.00602 \\
   \bottomrule
  \end{tabular}
\end{table*}

\newpage\phantom{blabla}
\newpage 
\subsection{Additional evaluations of the proposed method}

\subsubsection{Inversion error of individual facial expressions: comparisons across different training procedures and loss ablation}

We show the inversion quality for individual facial expressions using different methods (ReStyle, Ours w/o $\mathcal{L}_{{l2}_{eye}}$, and Ours) in \tabref{appendix-error}. The improved inversion quality of facial images featuring eye movements, such as \lookup{}, \lookdown{}, \lookleft{}, and, \lookright{}, in our method demonstrates the importance of incorporating the L2 loss for eyes $\mathcal{L}_{{l2}_{eye}}$ to improve the latent representations for macaque faces.

\subsubsection{Editing macaque faces via style mixing} 

\begin{figure*}[t]
  \centering
  \includegraphics[clip,width=\linewidth]{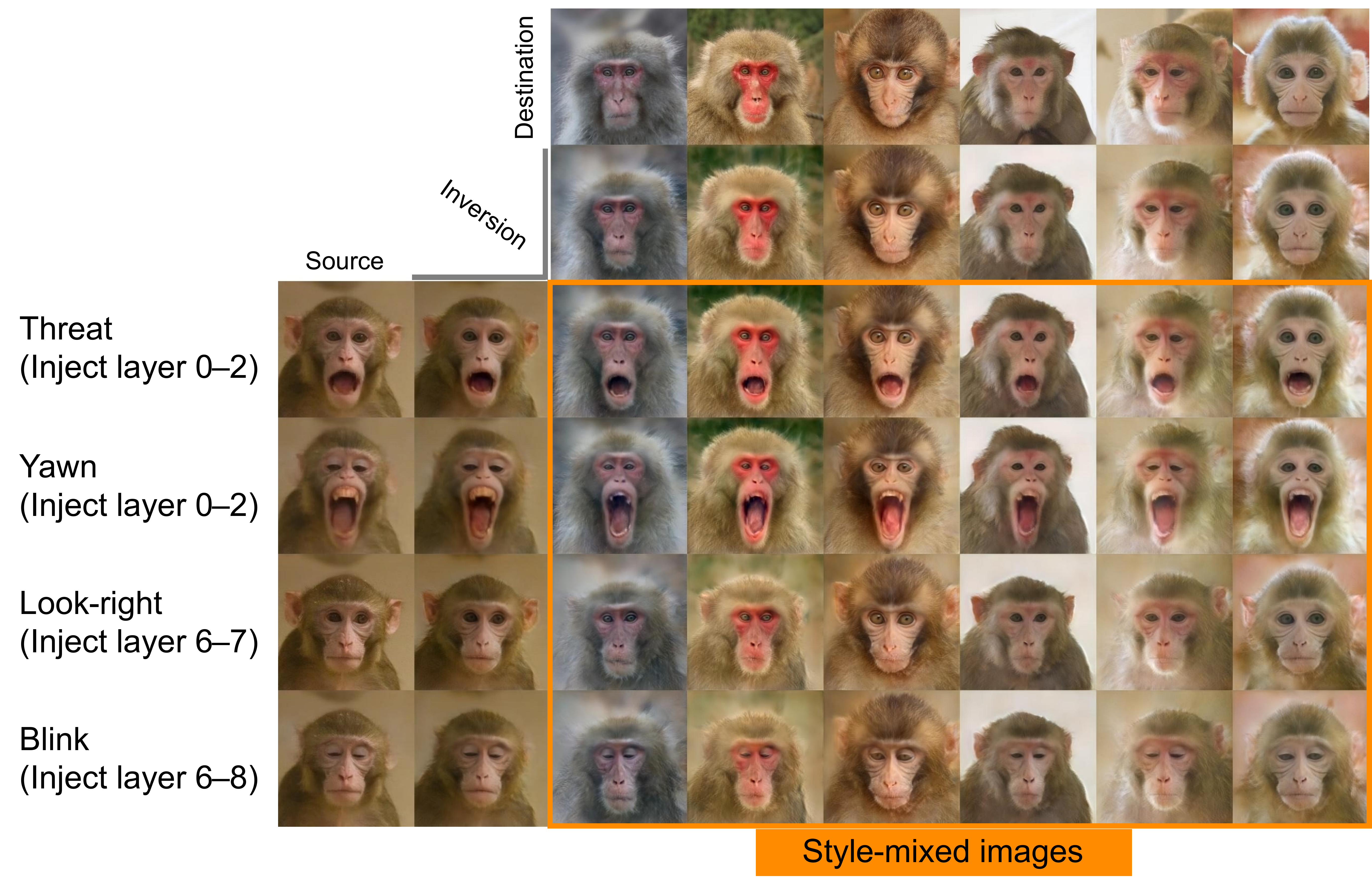}
  \caption{Results of the style mixing task. The results showcase the style mixing task using the model trained by our method on test images. The top row features the destination images and the second row displays their inverted versions. The first and second columns present the source images with different expressions and their inverted images, respectively. By injecting the latent codes of the source images into those of the destination images, the generated images (highlighted within the red contour) depict the macaque individuals of the destination images exhibiting the expressions of the source images.}
  \label{result-mix}
\end{figure*}

Style mixing is an image manipulation technique used to qualitatively demonstrate the disentanglement of the latent representations learned by the StyleGAN2 models~\cite{Karras2021a}. The process begins by selecting source and destination images, and then calculates their latent codes using the trained StyleGAN2 encoder. The latent codes from specific layers of the source image are injected into those of the destination image. The mixed latent representation is then used by the trained StyleGAN2 decoder to generate a new image. If the model has successfully disentangled facial expressions and identity in different layers, the generated image should display the facial expression of the source while maintaining the identity of the destination image. As shown in Fig.~\ref{result-mix}, we injected style parameters from layers 0--2, 0--2, 6--7, and 6--8 into those of the destination images when using \threat{}, \yawn{}, \lookright{}, and \blink{} expressions as source images, respectively.

Our method successfully transferred highly articulated expressions, such as \yawn{}, from the source image to the macaque individual in the destination image by selecting the latent code from specific layers of the StyleGAN2 model. This also demonstrates that semantically interpretable attributes in the images were represented hierarchically as the style parameter's values across the layers of the trained StyleGAN2 model. Eye movements could also be injected into other images, as shown in the images on the second-bottom row, which is important for controlling the direction of gaze and attention in macaque images. These results demonstrate that the StyleGAN2 model trained using our method had the capability to edit a range of expressions, from subtle simple facial movements, such as eye movements, to more conspicuous and complex expressions, such as \yawn{}. 





\begin{figure*}[t]
  \centering
  \includegraphics[clip,width=\linewidth]{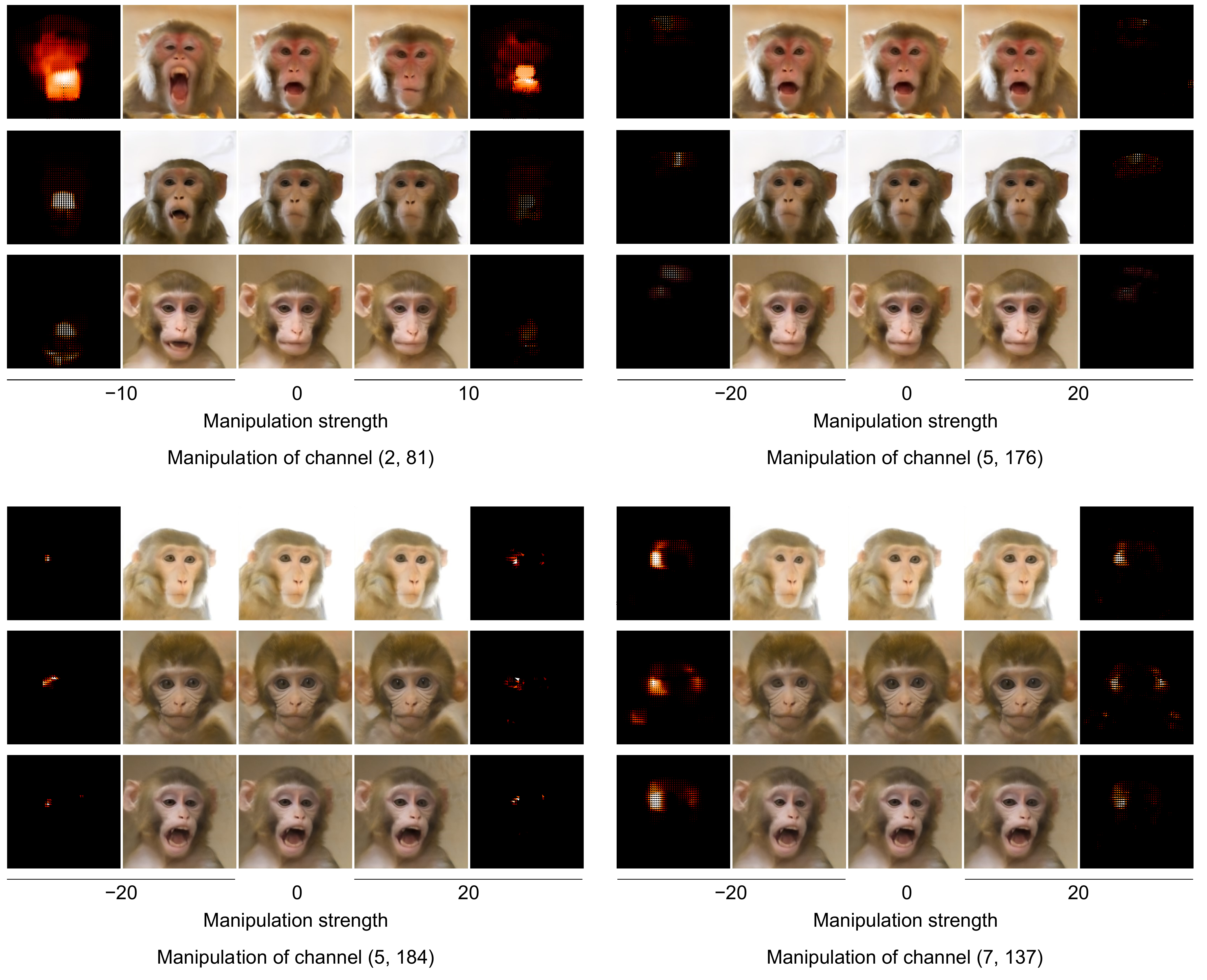}
  \caption{Examples of channel manipulation analysis in StyleSpace. This figure showcases four representative results of single-channel manipulation for three source images. Each panel illustrates the manipulation of mouth opening (channel no. 81 in layer 2), crown shape (channel no.176 in layer 5), eye direction (channel no. 184 in layer 5), and eye size (channel no.137 in layer 7). The scale of manipulation strength was manually adjusted for each channel to produce the results shown in this figure. Some features are exaggerated or deviated from the biological features to demonstrate the capabilities of the model. The monkey faces in the center column of each panel are the source images, whereas the images on the left and right display the edited results achieved by changing the target channel values in the negative and positive directions, respectively. The images on the far left and right depict the optical flow patterns calculated from the source and the manipulated images, highlighting the edited facial areas.}
  \label{fig:result-editchan}
\end{figure*}

\subsubsection{Manual inspection of the StyleSpace channels}
Another advantage of the latent exploration in the StyleSpace is that we can extract unlabeled attribute changes, which is useful for editing images. In Fig.~\ref{fig:result-editchan}, to qualitatively explore how individual channels contribute to the representation of specific visual features or actions, we manipulated individual channels and examined the differences between unmanipulated and manipulated macaque face images.
Manual inspection of the inversion results after modifying individual channels' outputs demonstrated that some channels were dedicated 
to controlling specific facial attributes, such as opening the mouth, the shape of the top of the head, eye direction, and eye size.

\subsubsection{Top 5 channels in StyleSpace activated by individual facial expressions}

The top channels for each facial expressions determined by StyleSpace analysis are shown in \tabref{appendix-chan}.
Expressions incorporating mouth opening movements were mainly represented in the course layers, such as layers 1 and 2.
By contrast, expressions related to eye movements, especially for \blink{}, \lookdown{}, and \lookright{}, were mainly represented in medium layers, such as layers 6, 7, and 8. 

\begin{table*}[t]
  \centering
  \caption{Top 5 channels for facial expression types in StyleSpace. 
  Parameters and channels in layer 0, which mainly represent very coarse image features that are not directly related to the facial features, were excluded from this analysis.}
  \label{tab:appendix-chan}
  \begin{tabular}{ll}
   \toprule
   Facial expression & (Layer, Channel, Rank) \\
   \midrule
   \baredteeth{} & (1, 226, 1), (1, 130, 2), (2, 81, 3), (1, 312, 4), (1, 187, 6) \\
   \bark{}       & (1, 130, 2), (1, 312, 3), (1, 226, 4), (1, 39, 7), (1, 187, 9) \\
   \blink{}      & (2, 355, 10), (2, 156, 13), (8, 56, 19), (7, 467, 25), (8, 135, 32) \\
   \browraise{}  & (5, 176, 2), (5, 272, 5), (3, 2, 8), (1, 428, 10), (1, 195, 14) \\
   \chewing{}    & (1, 130, 1), (1, 226, 2), (1, 312, 3), (1, 115, 5), (1, 357, 6) \\
   \coo{}        & (1, 302, 1), (1, 312, 2), (1, 130, 3), (1, 115, 4), (1, 300, 5) \\
   \lipsmack{}   & (11, 61, 1), (11, 2, 3), (12, 44, 4), (8, 237, 5), (12, 3, 6) \\
   \scream{}     & (1, 130, 5), (1, 226, 7), (1, 411, 8), (1, 141, 10), (1, 134, 11) \\
   \threat{}     & (1, 130, 2), (1, 492, 5), (1, 302, 6), (1, 411, 7), (1, 403, 8) \\
   \tongueprotrude{} & (6, 495, 1), (7, 163, 2), (2, 475, 3), (5, 84, 4), (12, 26, 5) \\
   \yawn{}       & (2, 81, 21), (2, 347, 42), (2, 246, 44), (2, 72, 45), (2, 144, 51) \\
   \lookup{}     & (1, 226, 4), (1, 431, 13), (1, 32, 14), (2, 216, 16), (1, 446, 22) \\
   \lookdown{}   & (6, 156, 7), (2, 355, 10), (2, 475, 11), (2, 17, 12), (2, 233, 17) \\
   \lookleft{}   & (2, 148, 1), (3, 128, 2), (2, 478, 3), (4, 356, 4), (1, 467, 7) \\
   \lookright{}  & (2, 277, 1), (6, 239, 2), (2, 469, 3), (7, 429, 4), (2, 360, 5) \\
   \tongueshow{} & (2, 31, 1), (2, 66, 2), (6, 413, 3), (2, 56, 4), (2, 314, 5) \\
   \bottomrule
  \end{tabular}
\end{table*}


\subsubsection{Morphing between different macaque individuals' images using latent representation}

The trained ReStyle encoder can convert macaque face images from different individuals into latent codes 
defined by style parameters. By interpolating the obtained latent codes and decoding 
from these interpolated parameters, we can generate morphing images between selected individuals as shown in \figref{appendix-morph}. 
These generated images appear photorealistic and can even resemble novel individuals. 
\begin{figure*}[t]
  \centering
  \includegraphics[clip,width=\linewidth]{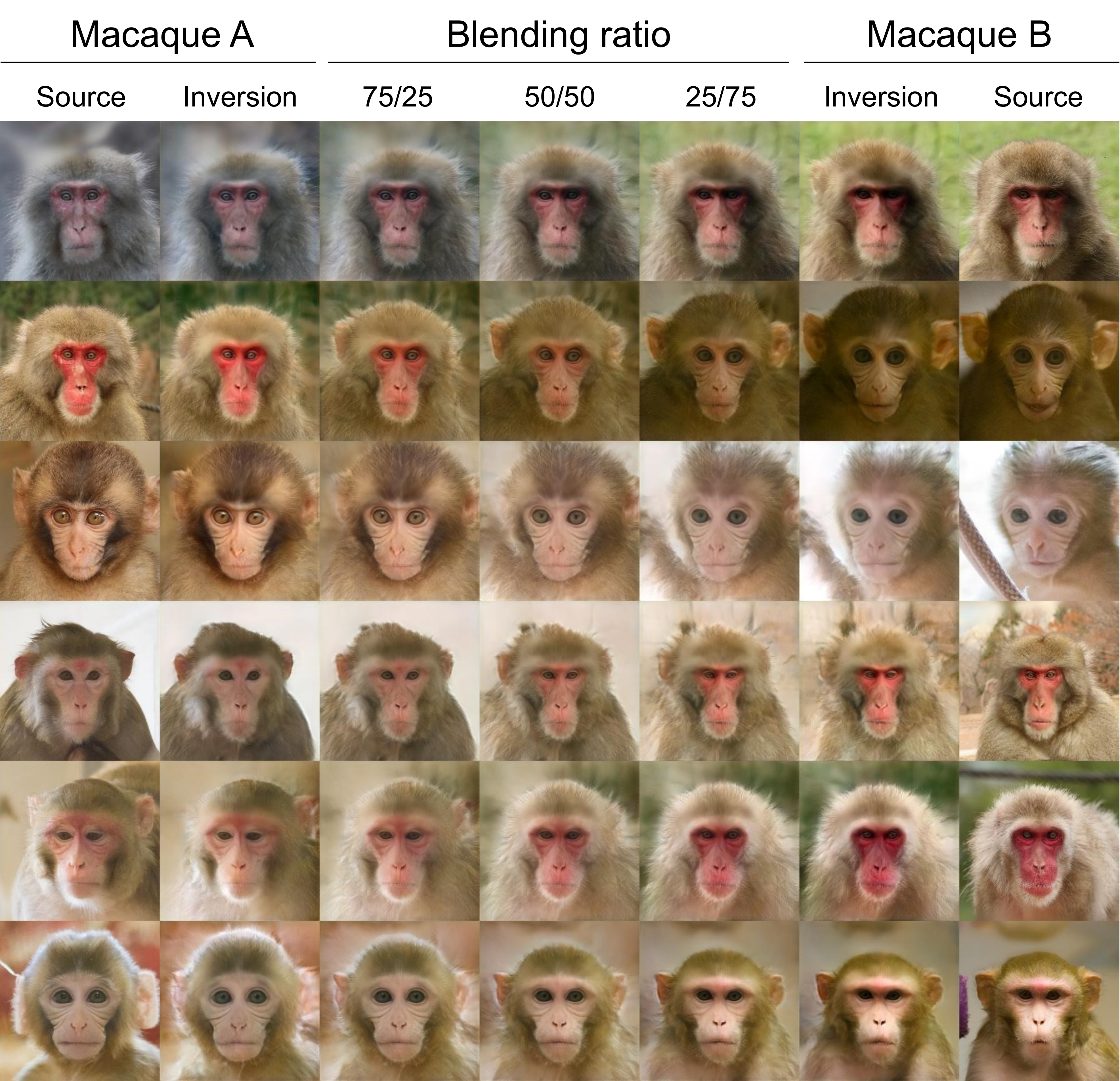}
  \caption{Examples of morphed face image generation using the linear interpolation of latent codes between faces of different macaque individuals. The images in the left and right columns are the original source macaque face images, whereas the second-left and second-right columns show the inverted images of the source images. The intermediate columns display images generated by linearly interpolating the latent codes of the two inverted images with blending ratios of 75/25\%, 50/50\%, and 25/75\%.}
  \label{fig:appendix-morph}
\end{figure*}

\end{document}